\documentclass{article}

\usepackage{microtype}
\usepackage{graphicx}
\usepackage{subcaption}
\usepackage{booktabs} % for professional tables

\usepackage{hyperref}

\usepackage{multirow}
\usepackage{makecell}
\usepackage{siunitx}
\usepackage{float}
\usepackage{microtype}
\usepackage{colortbl} 
\usepackage{xcolor}
\usepackage{enumitem}

\usepackage[workshop]{icml2026}

\usepackage{amsmath}
\usepackage{amssymb}
\usepackage{mathtools}
\usepackage{amsthm}

\usepackage[capitalize,noabbrev]{cleveref}

\theoremstyle{plain}

\theoremstyle{definition}

\theoremstyle{remark}

\usepackage[textsize=tiny]{todonotes}

\icmltitlerunning{Screening Drug Repositioning via Pretrained Representations}

\begin{document}

\twocolumn[
  \icmltitle{Pretrained Medical Representations for the Practical Screening of\\Drug Repositioning Candidates}



  \icmlsetsymbol{equal}{*}

  \begin{icmlauthorlist}
    \icmlauthor{Yuhei Fujioka}{cancerscan}
    \icmlauthor{Daitaro Misawa}{cancerscan}
    \icmlauthor{Shingo Fukuma}{kyoto-univ,hiroshima-univ}
  \end{icmlauthorlist}

  \icmlaffiliation{cancerscan}{Cancerscan Inc., Tokyo, Japan}
  \icmlaffiliation{kyoto-univ}{Kyoto University, Graduate School of Medicine, Kyoto, Japan}
  \icmlaffiliation{hiroshima-univ}{Hiroshima University, Graduate School of Biomedical and Health Sciences, Hiroshima, Japan}

  \icmlcorrespondingauthor{Yuhei Fujioka}{y.fujioka@cancerscan.jp}

  \icmlkeywords{Representation Learning, Healthcare, Drug Repositioning, Self-Supervised Learning, Hypothesis Generation}

  \vskip 0.3in
]



\printAffiliationsAndNotice{}  % no special notice (required even if empty)

\begin{abstract}
Representation learning from medical code sequences in electronic health records and medical claims data has been successful in various clinical applications, such as those regarding disease prediction. However, significant challenges remain in extending this approach to the discovery of scientific hypotheses. One reason is that many existing BERT-based models fail to adequately capture the hierarchical structure of medical codes and the complex interactions between diagnoses and treatments. 
To address these limitations, we propose a new unified pre-training framework that explicitly integrates hierarchical sub-token aggregation, partial masking, and cross-reference mechanisms. The proposed model consistently outperformed existing methods on both pre-training objectives and downstream clinical event prediction tasks, including the onset of dementia and hospitalization. 
We also conducted an \textit{in silico} drug repositioning case study targeting Alzheimer’s disease. In the hypothesis generation step, our approach successfully rediscovered known promising drugs (e.g., pitavastatin) in a data-driven manner without relying on such external knowledge sources as the literature. Subsequently, in the hypothesis prioritization step, we introduced a Task-Adaptive Representation Approach to alleviate the over-encoding of historical prescription information within diagnostic vectors, enabling the robust prioritization of generated hypotheses. This study establishes an exploratory screening workflow for hypothesis generation and prioritization based on observational associations. Importantly, this framework is not intended to provide causal evidence, but rather to identify promising candidates for subsequent rigorous causal inference. Overall, this study demonstrates that domain-informed representation learning combined with task-adaptive representation control can enable a practical hypothesis discovery workflow, beyond a single application domain.
\end{abstract}

\section{Introduction}
\subsection{Background}
Medical codes in claims data and electronic health records, such as diagnoses, procedures, and prescribed medications, comprise a rich source of clinical information. Machine learning methods leveraging these data have been applied to various medical tasks, including disease prediction \cite{Razavian2015-vo, Walsh2019-cm}, medication recommendation \cite{ijcai2019p825}, and drug repositioning \cite{zang2023high, lee2025deep}. Recently, there has been growing interest in applying BERT \cite{devlin-etal-2019-bert} to structured medical code sequences. However, directly adopting the standard BERT pre-training approach presents several challenges. First, it often fails to capture the hierarchical structure inherent in medical coding systems (e.g., ICD-10) and the complex relationships between diagnoses and treatments (procedures and prescribed drugs) \cite{fujioka2024in}. 
Second, standard masked language modeling (MLM) is inefficient for fine-grained medical codes. For example, there are approximately 50 distinct codes within the single category of diabetes. Requiring the model to predict such highly specific codes can both increase training difficulty and reduce generalizability. Consequently, previous studies have not adequately addressed these challenges, thereby failing to effectively use the  information contained in medical code data.

\subsection{Task Definition}
 We developed a representation learning model using medical claims data and demographic information, and evaluated its generalization ability and applicability to hypothesis discovery through three tasks: (i) a pre-training task in which the model learns representations by predicting MLM-targeted diagnosis sub-tokens, (ii) clinical event prediction tasks for the onset of dementia and hospitalization to assess generalization; and (iii) a drug repositioning task that uses vector representations generated by the pretrained model to select candidate drugs showing protective associations against Alzheimer’s disease (AD) (hypothesis generation) and prioritize them based on signal robustness (hypothesis prioritization).

\subsection{Challenges}\label{Challenges}
Applying standard BERT models to medical code sequences entails three major challenges: (i) the absence of hierarchical information modeling, (ii) the standard MLM’s inefficiency for fine-grained medical codes, and (iii) the absence of explicit modeling frameworks for diagnoses–treatment interactions. In particular, self-attention treats diagnoses and treatments uniformly, which risks obscuring clinically important relationships. Moreover, there exists a tension between representations encoding rich clinical information and those suitable for hypothesis screening or confirmatory inference. Representations that encode historical prescription information can enhance signal detection (hypothesis generation) by leveraging rich clinical information, yet undermine robustness in validation-oriented analyses such as propensity score–based hypothesis prioritization.

\subsection{Contributions} 
This study makes four main contributions through representation learning methods that explicitly account for the characteristics of medical code sequences: (i) we propose a new unified pre-training framework that integrates Hierarchical Sub-token Aggregation (HSA) to depict the hierarchical structure of medical codes, Partial Masking (PM) to improve training efficiency, and a Cross-Reference (CR) mechanism to model the diagnosis–treatment interaction, thereby improving the quality of representations; (ii) we demonstrate the effectiveness of the proposed model in clinical event prediction tasks, achieving superior performance than existing benchmarks for predicting the onset of dementia and hospitalization.; (iii) we show that the pretrained representations enable data-driven hypotheses generation by rediscovering promising candidate drugs for AD (e.g., pitavastatin) without relying on external knowledge sources such as the literature; and (iv) we introduce a Task-Adaptive Representation Approach (TARA) that resolves the tension between detection-oriented and validation-oriented representations, enabling robust hypothesis prioritization and establishing an effective screening workflow prior to costly causal inference.

\section{Related Work}
Shang et al. \cite{ijcai2019p825} demonstrated that learning the hierarchical structure of medical codes using graph neural networks could improve the performance of medication recommendation models. We incorporated this insight into our model through hierarchical sub-token aggregation within a Transformer-based architecture. Existing pretrained models such as BEHRT \cite{Li2020-op}, MED-BERT \cite{Rasmy2021-om}, and ExBEHRT \cite{10.1007/978-3-031-39539-0_7} have achieved notable success in representation learning for medical code sequences. However, the standard MLM’s above-mentioned inefficiency for fine-grained codes and modeling complex relationships between diagnostic and treatment codes (procedures and prescriptions) remains. Recently, drug repositioning using real-world data has attracted increasing attention \cite{zang2023high, lee2025deep}. While causal inference-based methods are powerful, they are computationally expensive and difficult to scale to a large number of candidate drugs. Consequently, there is a growing need for efficient screening methods to identify promising candidates before estimating causal effects. This study presents a practical screening framework for drug repositioning that leverages representations learned from pre-training, reducing overall analytical cost.

\section{Methods}
\begin{figure*}[t]
    \centering
    \includegraphics[width=0.8\linewidth, height=4.5cm, keepaspectratio]{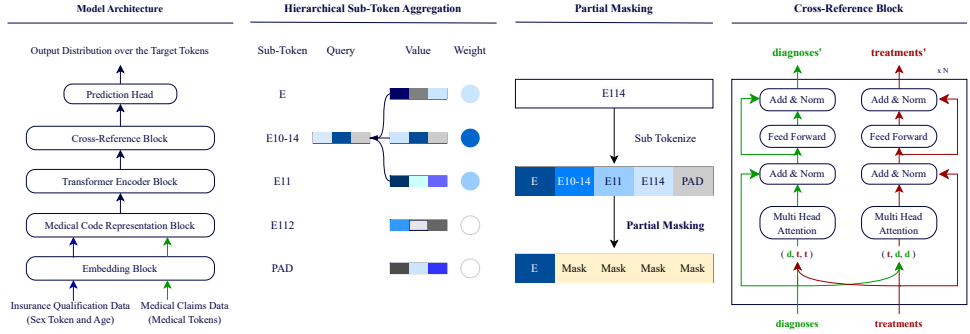}
    \caption{\textbf{Overview of our proposed architecture and methods.}}
    \label{fig_our_model_archi}
\end{figure*}
\subsection{Pre-training}
Figure \ref{fig_our_model_archi} presents an overview of the proposed methods.
\subsubsection{Hierarchical Sub-Token Aggregation}
To leverage the hierarchical structure of medical tokens, we divided each token into five sub-tokens. Each sub-token was embedded in a vector representation. The model was trained to use the hierarchical relationships among the five sub-tokens using a Transformer encoder.
\subsubsection{Partial Masking}
Fifteen percent of the tokens from each sequence are randomly selected as targets for MLM. If a sequence is too short and no token is selected, one token is randomly chosen to ensure at least one MLM target. Each selected token is then decomposed into five sub-tokens according to its hierarchical structure. For these, one of three operations is applied: partial masking (PM), full masking, or no change, with respective probabilities of 80\%, 10\%, and 10\%. In PM, the first sub-token is retained, while the second through fifth sub-tokens are masked. In full masking, all five sub-tokens are masked. In the no-change case, all sub-tokens are kept unchanged. The model is trained to predict the third sub-token, which corresponds to the hierarchy’s middle level. Existing benchmark models follow the standard BERT masking strategy. An overview of the masking strategy is provided in Appendix \ref{appendix masking strategy}. 
\subsection{Model}
\subsubsection{Our Pre-training Model}\label{our pre-training model}
Our pre-training model was designed to address both the hierarchical structure of medical tokens and the complex relationships between diagnoses and treatments. The model architecture (Figure \ref{fig_our_model_archi}) consists of five blocks designed to process monthly medical claims and insurance qualification data (for detailed descriptions of each block, see Appendix \ref{appendix archi supplement}). \textbf{(i) Embedding Block:} Each medical token (diagnosis, procedure, and prescription) is decomposed into five sub-tokens based on its hierarchical structure, each of which is then embedded into a vector. The sex token is embedded in the same manner, while age is transformed into a vector using a single linear layer. \textbf{(ii) The Medical Code Representation Block:} For each type of medical sub-token sequence, independent Transformer encoders designed to perform hierarchical sub-token aggregation (HSA) are applied respectively. Each sub-token representation is updated using information from itself and neighboring sub-tokens within the same medical token. The updated sub-token representations belonging to the same token are then summed to form diagnosis, procedure, and prescription token sequences. For the diagnosis sequence, sex and age vectors are added to each representation. \textbf{(iii) Transformer Encoder Block:} Procedure and prescription token vectors are concatenated to form treatment vectors. The diagnosis and treatment sequences are then processed independently by separate Transformer encoders. \textbf{(iv) Cross-Reference (CR) Block:} A bidirectional cross-attention mechanism captures the structured interactions between diagnoses and treatments and refines their vector representations. \textbf{(v) Prediction Head:} The model predicts the MLM-targeted diagnosis sub-token based on the output of the final hidden states of diagnosis tokens from the CR block.
\subsubsection{Our Fine-Tuning Model}
 To assess the quality and generalizability of the pretrained representations, we developed a fine-tuning model for downstream clinical event prediction tasks, including the onset of dementia and hospitalization. Detailed descriptions of the model is provided in Appendix \ref{appendix fine-tuning model}.

\subsection{Benchmark Models}
We compared our approach with four baseline models: (i) \textbf{BEHRT} and (ii) \textbf{MED-BERT}, which are representative self-attention-based models for medical code sequences; \textbf{(iii) Our Self-Attention Only Model (Our (SA))}, an ablation model designed to evaluate the effectiveness of the cross-reference mechanism by adopting our proposed methods (HSA and PM) within a standard self-attention architecture ; and \textbf{(iv) Light Gradient Boosting Machine (LGBM)} \cite{NIPS2017_6449f44a}, a gradient boosting framework using tabular input features. Detailed architectures and input configurations are provided in Appendices \ref{appendix hyperparameter} and \ref{appendix benchmark models}.

\subsection{Task-Adaptive Representation Approach}\label{TARA}
 The TARA is an inference-time representation control strategy designed to address the varying suitability of diagnostic representations for hypothesis generation and  prioritization. As noted in Section \ref{Challenges}, diagnostic representations that incorporate treatment history are useful for generating AD risk profiles that reflect broader clinical context, making them important to hypothesis generation. However, during hypothesis prioritization, such representations can over-encode the propensity for prescription assignment, potentially affecting the estimation of propensity scores and subsequent prioritization analyses in an adverse manner. TARA addresses this issue by controlling input tokens based on the analysis stage (i.e., candidate drug selection or prioritizing selected candidates). Specifically, during representation inference for hypothesis prioritization, prescription tokens corresponding to the target drug and related medications within the same second level of the Anatomical Therapeutic Chemical (ATC) classification system are masked when generating diagnosis vectors. This control over vector representations reduces access to historical prescription information for the target drug and its pharmacological neighbors, thereby improving overlap in propensity-score distributions between the prescribed and control groups.
\subsection{Drug Repositioning Procedures}
We designed a drug repositioning analysis framework for AD using real-world data.
\subsubsection{A Temporally Ordered Observational Study Design}
Our drug repositioning analysis followed a temporally ordered observational design. Specifically, claims data from January to December 2017 were used to extract diagnostic vectors from the pretrained model; these served as covariates for estimating prognostic scores (AD onset risk) and propensity scores. The period from January to December 2018 was defined as the exposure assessment window. For each drug, regular prescription status was defined as having prescriptions recorded in at least six monthly claims during this period, and this drug-specific prescription status was used as the exposure indicator. The follow-up period spanned from January 2019 to December 2020, during which the occurrence of AD onset was assessed using ICD-10 codes F00 and G30.
\subsubsection{Candidate Drug Selection}\label{candidate selection method}
Candidate drugs for AD prevention were selected through a two-stage procedure. \textbf{(i) Risk Matching Based on Prognostic Scores:} To adjust for baseline disease risk, we performed matching based on the predicted probability of AD onset two years later (prognostic score). Prognostic scores were estimated using logistic regression, Covariate Balancing Propensity Score (CBPS) \cite{imai2014cbps}, and LGBM, with sex, age, and a diagnosis vector generated from the pretrained model as covariates. We defined the diagnosis vector as the sum of the final transformer layer outputs over one year of diagnosis token sequences. Notably, the prognostic scores are learned from the entire study population without distinguishing treatment status (i.e., regular use of the target drug) for computational efficiency. As a consequence, treatment effects may be partially absorbed into the risk scores (prognostic scores) \cite{hansen2008prognostic}, potentially yielding conservative estimates in subsequent analyses and biasing odds ratios (ORs) toward 1.0. \textbf{(ii) Hypothesis Generation via Candidate Selection:} After prognostic score matching, logistic regression was applied to assess the association between regular drug prescription and AD onset. Candidate drugs were selected based on screening criteria using both estimated ORs and \textit{p}-values.
\subsubsection{Prioritizing Selected Candidates} 
To prioritize the selected candidates for further investigation, we adopted an analysis design that leverages informative diagnosis representations while preventing prescription-related information encoded in the vectors from leaking into the estimation of propensity scores via TARA. \textbf{(i) Propensity Score Matching:} Propensity scores were estimated using sex, age, and diagnosis vectors as covariates. We employed CBPS for propensity score estimation, as it achieved the most favorable covariate balance in the prognostic score matching. \textbf{(ii) Association Analysis and Prioritization:} Following matching, logistic regression was performed to estimate the association between regular prescription of each candidate drug and the onset of AD after two years. Finally, multiple testing correction was performed on the estimated associations, and the candidates were prioritized based on the results.

\subsection{Input Features}
We used medical claims data and insurance qualification data recorded from January to December 2017 as inputs to both the pre-training and fine-tuning models. We also used medical claims data recorded from January 2017 to December 2018, together with insurance qualification data, to construct input features for score estimation and statistical analyses in the drug repositioning task. Based on these data, we constructed two types of input features depending on the model architecture and task type: monthly input features, consisting of medical codes (diagnosis, procedure, and prescription codes) from monthly claims data along with demographic information (sex and age) from insurance qualification data; and annual input features, derived from processing these monthly sequences into concatenated, stacked, or tabular formats. A detailed description of each feature type and the construction process is provided in Appendix \ref{appendix_input_features}.

\subsection{Training}
We used the AdamW optimizer for training \cite{loshchilov2018decoupled}. Cross-entropy loss was applied during pre-training, while Class-Balanced Focal Loss \cite{Cui_2019_CVPR} was used during fine-tuning to address label imbalance. LGBM Hyperparameter tuning was performed using Optuna \cite{10.1145/3292500.3330701}.

\section{Experiments}
\subsection{Experimental Setting}
To evaluate the effectiveness of the three proposed components—\hspace{0pt}Hierarchical Sub-token Aggregation (HSA), Partial Masking (PM), and Cross-Reference (CR)—in both pre-training and downstream tasks, we compared models incorporating the proposed methods with four baselines: BEHRT, MED-BERT, Our (SA), and LGBM.

\subsubsection{Pre-training Task}
In the pre-training task, all models were trained to predict the third-level classification of diagnosis tokens selected as targets in MLM. We first compared the model integrating all proposed methods against benchmark models to assess the overall effectiveness of the proposed framework. Next, to disentangle the contributions of HSA, PM, and architectural differences, we conducted an ablation study using the CR-based architecture and an SA-only architecture. The dataset was divided into a ratio of 8:1:1 for the training, validation, and test sets, using KFold. The models were trained for 25 epochs, and the checkpoint with the lowest validation loss was used for evaluation. Moreover, we compared learning curves across proposed methods and examined the impact of HSA and PM on prediction performance for diagnosis tokens with different occurrence frequencies, distinguishing between high- and low-frequency tokens.
\subsubsection{Visualization of the Disease Embedding} To assess the quality of representations learned during pre-training, diagnosis vectors extracted from the final transformer layer of pretrained models (Our (CR), BEHRT, and MED-BERT) were projected into two dimensions using t-SNE \cite{JMLR:v9:vandermaaten08a}. From approximately 500,000 diagnosis tokens in the pre-training test set, 50,000 tokens were randomly sampled to examine whether diagnoses belonging to the same disease category formed coherent clusters in the embedding space.

\subsubsection{Clinical Event Prediction Tasks}\label{experimental setting clinical event prediction tasks}
We also conducted clinical event prediction tasks for dementia onset and hospitalization to evaluate the generalization ability of the pretrained representations. The detailed experimental setting is provided in 
Appendix \ref{appendix_clinical_predicton_task_experimental_setting}.

\subsubsection{Candidate Selection for Drug Repositioning}
We conducted a case study focusing on AD. First, prognostic scores were estimated using logistic regression, LGBM, and CBPS, with diagnosis vectors from pretrained models (Our (CR) trained with both HSA and PM, BEHRT, and MED-BERT), sex, and age as covariates. Nearest-neighbor matching was implemented using a 1:3 ratio and a caliper width of 0.2 standard deviations, targeting the Average Treatment Effect on the Treated (ATT). CBPS was selected to extract case–control pairs based on covariate balance assessed by standardized mean differences (SMDs). Logistic regression was applied to this case-control group to analyze the association between regular drug prescription and AD onset. Following previous studies, only drugs with at least 500 regular users were included \cite{zang2023high, lee2025deep}, and drugs satisfying screening criteria (OR $<$ 0.9 and \textit{p} $<$ 0.05) were selected as candidates. Here, the \textit{p}-values were used for exploratory screening rather than confirmatory inference. To assess the existing evidence, we conducted a comprehensive literature search using Gemini Deep Research, followed by a manual review of previously reported evidence linking the candidate drugs to AD.

\subsubsection{Prioritization of Selected Candidates}
We prioritized the selected candidates based on the strength of association between regular prescription and AD onset. This analysis sought to prioritize hypotheses generated from observational data for further investigation, rather than to establish causal effects. Candidate selection and this prioritization analysis were conducted on the same drug repositioning cohort; therefore, the reported \textit{p}-values were used solely for hypothesis generation and prioritization, not for confirmatory inference. Propensity scores were estimated using CBPS with sex, age, and diagnosis vectors from the pretrained models (Our (CR) trained with both HSA and PM, BEHRT, and MED-BERT) as covariates. Our (CR) used the diagnosis vectors generated using TARA (Section \ref{TARA}), whereas TARA was not applied to BEHRT or MED-BERT, as their inputs consisted solely of diagnosis tokens. Nearest-neighbor matching was then performed with estimand ATT, a 1:2 matching ratio, and a caliper width of 0.2 standard deviations. Logistic regression was applied to the matched cohorts, with regular prescription status as the explanatory variable and AD onset within two years as the outcome variable, to estimate ORs and \textit{p}-values. Finally, consistent with previous studies \cite{lee2025deep}, multiple testing correction was performed across candidate drugs with a false discovery rate (FDR) of 0.05, and FDR-adjusted \textit{p}-values below 0.05 were prioritized as candidate hypotheses.

\subsection{DataSets}\label{datasets}
We used Japanese medical claims and insurance qualification data. The claim data were partitioned into a pre-training/drug repositioning group (80\%, \textit{n}=4,405,316) and a clinical event prediction group (20\%, \textit{n=}1,101,329) to ensure fair evaluation and prevent data leakage. Despite the full dataset of over 5 million individuals, we efficiently conducted experiments on sample subsets of tens of thousands under computational constraints. This design is intended to demonstrate the data efficiency of our methods and efficiently validate research hypotheses as a proof of concept, rather than to maximize predictive performance at scale. Further details are provided in Appendix \ref{appendix datasets}.

\begin{table*}[htbp]
  \caption{\textbf{Pre-training task results.} The column ”Proposed Method” indicates the applied approaches, where HSA means Hierarchical
Sub-token Aggregation and PM means Partial Masking.}
  \label{tab:combined}
  \centering
  \captionsetup[sub]{font=small,labelfont=bf,justification=centering,singlelinecheck=false}
  \renewcommand{\arraystretch}{0.8}
  \setlength{\tabcolsep}{2pt}
  \begin{subtable}[t]{0.35\linewidth}
    \caption{Pre-training task results}
    \centering
    \captionsetup{justification=centering,singlelinecheck=false}
    \begin{scriptsize}
      \resizebox{\linewidth}{!}{
        \begin{tabular}{cccccc}
            \toprule[0.4mm]
                \multirow[c]{3}{*}{Model} & \multirow[c]{3}{*}{Proposed Method} & & \multirow[c]{3}{*}{\makecell[c]{Accuracy}} & \multirow[c]{3}{*}{MCC} & \multirow[c]{3}{*}{\makecell[c]{Balanced\\Accuracy}} \\ 
                & & & & & \\
                 & & & & & \\  \midrule[0.1mm]\midrule[0.1mm]
                 \multirow[c]{3}{*}{Our (CR)} &  & & 0.3783 & 0.3656 & 0.2707\\
                 & HSA  & & 0.3913 & 0.3790 & 0.2961\\
                 & HSA + PM  & & \textbf{0.5867} & \textbf{0.5797} & \textbf{0.4710}\\ \midrule[0.1mm]
                 \multirow[c]{3}{*}{Our (SA)} & &  & 0.3570 & 0.3440 & 0.2479\\
                 & HSA & & 0.3867 & 0.3744 & 0.2952\\
                 & HSA + PM & & 0.5804 & 0.5732 & 0.4639 \\ \midrule[0.1mm]
                 BEHRT &  & & 0.1840 & 0.1676 & 0.1204\\ \midrule[0.1mm]
                 MED-BERT & & & 0.2553 & 0.2400 & 0.1091 \\
            \bottomrule[0.4mm]
        \end{tabular}
      }
    \end{scriptsize}
    \label{tab1:left}
  \end{subtable}%
  \hfill
  \begin{subfigure}[t]{0.32\linewidth}
    \centering
    \caption{Pre-training Learning Curve}
    \captionsetup{justification=centering,singlelinecheck=false}
    \includegraphics[
        height=0.63\linewidth,
        keepaspectratio
    ]{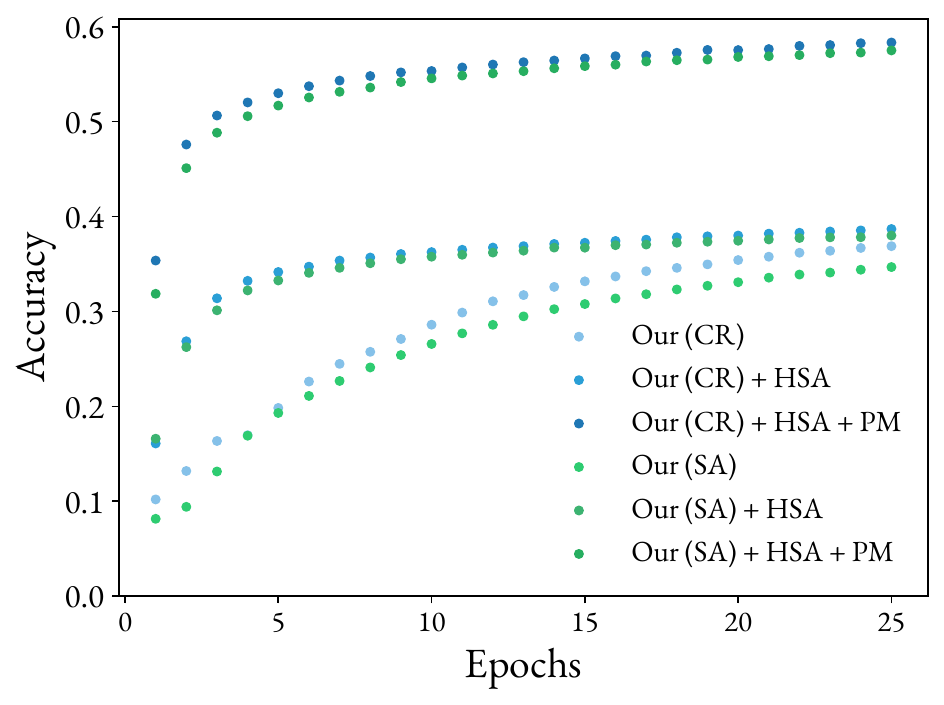}
    \label{fig1:right}
   \end{subfigure}%
   \hfill
  \begin{subtable}[t]{0.3\linewidth}
    \captionsetup{justification=raggedright,singlelinecheck=false}
    \caption{Mean MCC scores for frequent (top 10 frequent) and infrequent (the 141st–150th least frequent) diagnoses}
    \centering
    \begin{scriptsize}
      \resizebox{\linewidth}{!}{
        \begin{tabular}{ccccc}
            \toprule[0.4mm]
            \multirow{2}{*}{Model} & \multirow{2}{*}{\makecell[c]{Diagnosis \\Type}} & \multicolumn{3}{c}{Proposed Method}  \\\cmidrule[0.1mm]{3-5}
             & & & HSA & HSA + PM \\ \midrule[0.1mm]\midrule[0.1mm]
            \multirow{2}{*}{Our (CR)} & Frequent & 0.440 & 0.443 &  \textbf{0.700}\\
             & InFrequent & 0.219 & 0.291 & \textbf{0.373}\\ \bottomrule[0.4mm]
        \end{tabular}
      }
    \end{scriptsize}
    \label{tab1:right}
  \end{subtable}
\end{table*}
\subsection{Metrics}
To evaluate pre-training performance, we used Accuracy, MCC, and Balanced Accuracy---all of which are suitable for multiclass classification. For the clinical event prediction tasks, we used PR-AUC, MCC, and ROC-AUC to address class imbalance. MCC was computed using a fixed decision threshold of 0.5.
\section{Results and Discussion}
\subsection{Pre-training Task}\label{results_pre-training_task}
Table \ref{tab1:left} lists the evaluation results of the pre-training task. Introducing the proposed techniques---HSA and PM--consistently improved all evaluation metrics for both the Cross-Reference (CR) and Self-Attention (SA) models. Indeed, PM produced substantial improvements of over 50\% for most metrics compared to models without PM. Even without applying HSA and PM, the CR architecture outperformed both Our (SA) and such existing benchmarks as BEHRT and MED-BERT. Overall, the model integrating all three techniques (HSA, PM, and CR) achieved the best performance. Table \ref{fig1:right} presents the learning curves, illustrating the impact of HSA and PM on training efficiency. While HSA alone improved both convergence speed and accuracy, combining HSA with PM dramatically accelerated training. Compared to baseline models, the proposed approach achieved significantly higher accuracy with fewer training epochs. The analysis, stratified by diagnosis code frequency (Table \ref{tab1:right}), suggests that  complementary mechanisms may underlie the proposed methods. Introducing HSA alone improved the MCC score for low-frequency diagnoses (+7.2 points), likely due to the enhanced contextualization enabled by hierarchical information. In contrast, the combination of HSA and PM yielded substantial improvements for both high- (+26.0 points) and low-frequency (+15.4 points) diagnoses. This improvement can be attributed to PM revealing higher-level hierarchical information as hints, thereby efficiently narrowing prediction targets, particularly for high-frequency tokens. In sum, the complementary effects of PM—enhancing prediction for frequent classes—and HSA—mitigating the difficulty of predicting rare classes—effectively address the long-tail challenge inherent to medical code data and substantially improve overall predictive performance.

\subsection{Visualization of the disease embedding}
The t-SNE visualizations of the pretrained diagnosis vectors (provided in Appendix \ref{appendix disease embedding}) showed that the CR model equipped with HSA and PM produced clear clusters corresponding to disease categories. In contrast, the benchmark models failed to form such well-separated clusters. These results indicate that the proposed approach learns more discriminative and medically meaningful representations.

\subsection{Clinical Event Prediction Tasks}
Clinical event prediction results (Table \ref{results of clinical event prediction tasks}) show that Our (CR) with HSA and PM achieved the best performance for dementia onset, with a PR-AUC improvement of over 10\%. Conversely, improvements for hospitalization were limited, 
suggesting the model aligns with long-term dementia progression better than acute clinical events, reflecting differences in the temporal characteristics of these outcomes.

\begin{table*}[h]
\setcounter{table}{2} 
\centering
\caption{\textbf{Evidence Status of Candidate Drugs} (See Appendix \ref{appendix: summary of literature review} for the summary of the literature review) .}
\begin{small}
\label{table:evidence status}
\begin{tabular}{c p{3.7cm} p{5.3cm} p{4.7cm}}
\toprule
& \multicolumn{2}{c}{\textbf{Reported Protective Associations}} & \multicolumn{1}{c}{\multirow{2}{*}{\textbf{Not Reported}}}\\ 
\cmidrule(lr){2-3} 
\textbf{Model} & \multicolumn{1}{c}{\textbf{Human Studies}} & \multicolumn{1}{c}{\textbf{Preclinical or Indirect Evidence}} &  \\
\midrule
Our (CR) 
& Candesartan \cite{lundin2024association}, Pitavastatin \cite{westphal2025statin}, Alendronic Acid \cite{sing2025bisphosphonates}
& Vildagliptin \cite{ma2018vildagliptin}, Loxoprofen \cite{vom2025long}, Irbesartan and Amlodipine \cite{lundin2024association} 
& Mecobalamin, Platelet Aggregation Inhibitors excluding Heparin (PAIH) \\
\addlinespace
BEHRT 
& Candesartan, Pitavastatin 
& Vildagliptin, Loxoprofen, Ketoprofen \cite{vom2025long}, 
& Amlodipine, Mecobalamin, Adenosine, Other Ophthalmologicals\\
\addlinespace
MED-BERT 
& Candesartan, Pitavastatin 
& Vildagliptin, Teneligliptin \cite{wang2023teneligliptin}  
& Mecobalamin, PAIH, Pregabalinum, Other Ophthalmologicals \\
\bottomrule
\end{tabular}
\end{small}
\end{table*}
\subsection{Candidate Selection for Drug Repositioning
}\label{result_candidate_selection}

\begin{figure}[h]
    \centering
    \begin{minipage}[b]{0.5\linewidth}
        \centering
        \includegraphics[width=\linewidth]{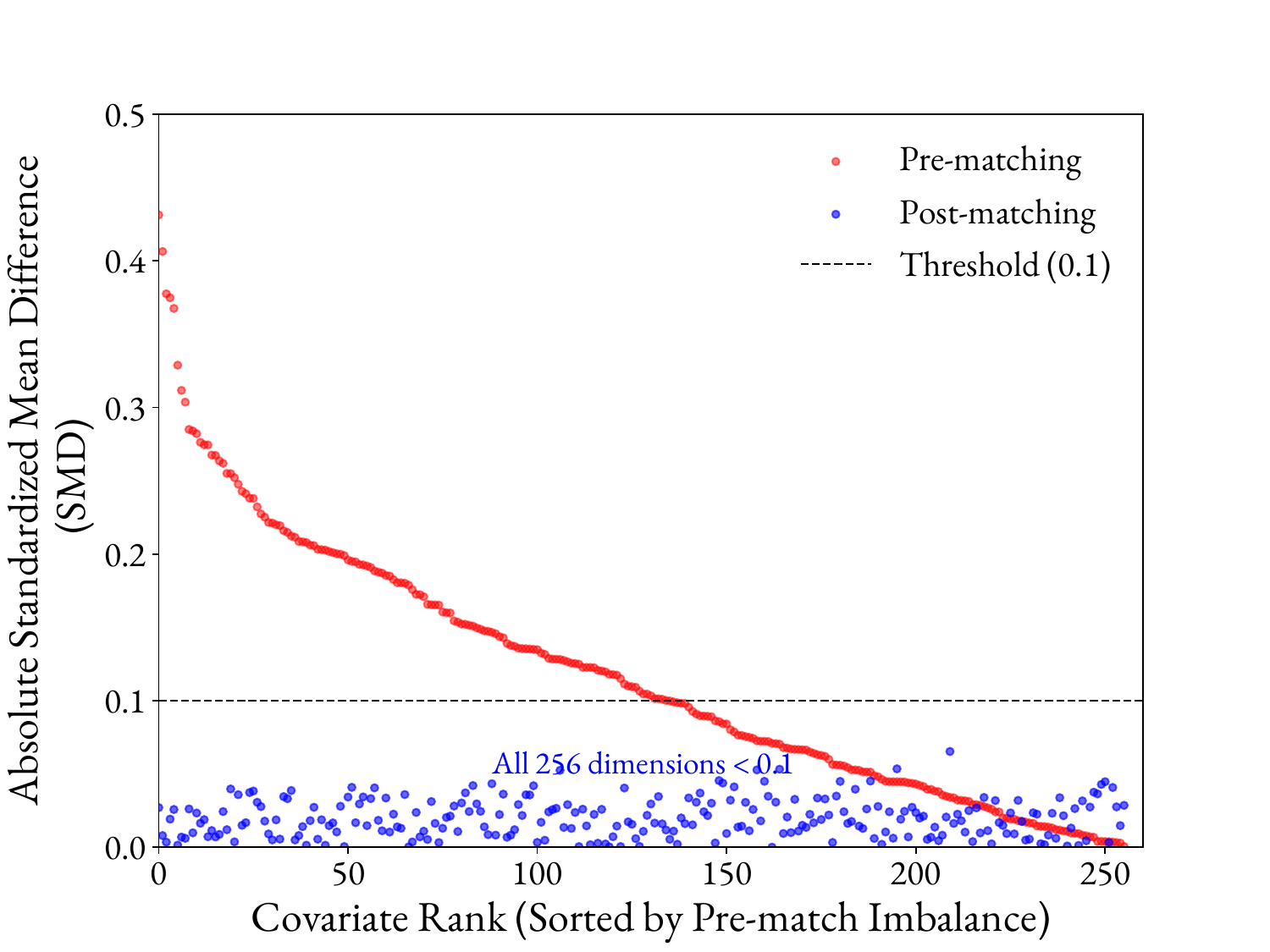}
        \subcaption{Covariate Balance} 
        \label{fig:prognostic_covariates_smd}
    \end{minipage}%
    \hfill
    \begin{minipage}[b]{0.5\linewidth}
        \centering
        \includegraphics[width=\linewidth]{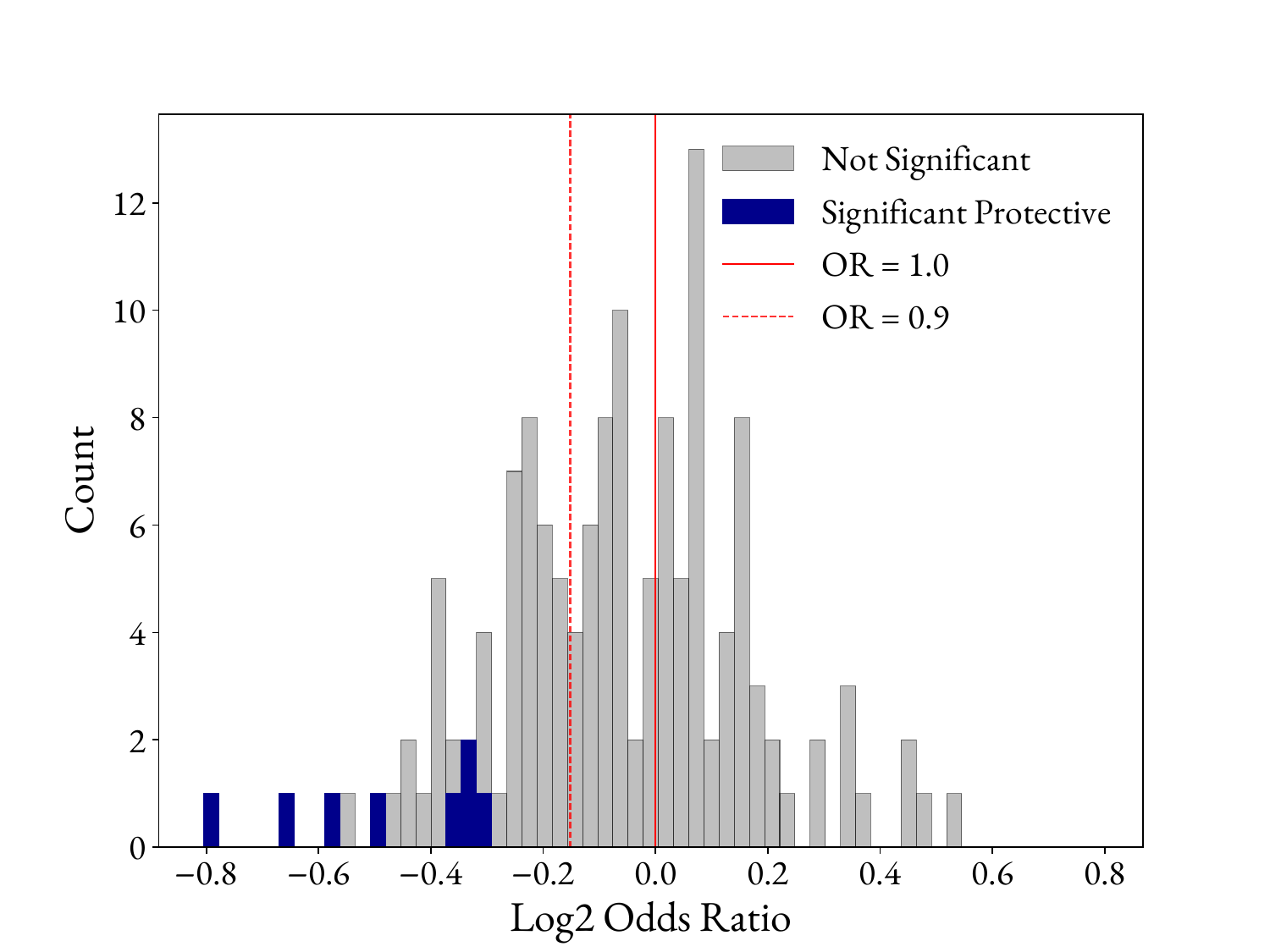}
        \subcaption{Distribution of Drug ORs}
        \label{fig:dist_estimated_ors}
    \end{minipage}
    \caption{\textbf{Performance of CBPS-based prognostic score matching and subsequent drug screening using Our (CR) representations.} (a) Covariate balance of diagnosis vectors before and after matching. All vector dimensions are well-balanced, falling below the 0.1 threshold (dashed line) after the matching procedure. (b) Distribution of estimated drug odds ratios (ORs).}
\end{figure}

Due to prognostic score matching, LGBM failed to sufficiently balance the covariates, whereas both logistic regression and CBPS achieved good balance (SMD $<$ 0.1) across all covariates (sex, age, and diagnosis vectors), irrespective of diagnosis vector representation used. Figure \ref{fig:prognostic_covariates_smd} shows the covariate balance achieved by CBPS-based prognostic matching using diagnosis vectors from our proposed model, visualized via SMDs across all 256 dimensions. Although logistic regression and CBPS performed comparably, we adopted the latter for subsequent analyses due to its theoretical robustness in optimizing covariate balance.
\begin{table}[H]
    \setcounter{table}{1} 
    \caption{\textbf{SMD of clinical severity indicators after prognostic score matching}}
    \centering
    \small
    \begin{tabular}{cccc}
        \toprule
        & Our (CR) & BEHRT & MED-BERT \\
        \midrule
        CCI & 0.014 & 0.014 & 0.002 \\
        NDPM$^1$ & 0.012 & 0.037  & 0.052 \\
        \bottomrule
    \end{tabular}
    \flushleft \footnotesize{1) Number of distinct prescribed medicines identified by ATC codes per year.}
    \label{table:smd of severity and healthcare utilization after prognostic matching}
\end{table}
Table \ref{table:smd of severity and healthcare utilization after prognostic matching} lists differences in clinical severity indicators, reflecting both the disease aspect (Charlson Comorbidity Index, CCI \cite{charlson1987new}) and the treatment aspect (number of distinct prescribed medicines per year), between cases (AD onset) and controls (no AD onset) after prognostic score matching. Across all models, SMDs for the indicators remained below 0.1, suggesting that the matched groups exhibited similar clinical severity. Taken together with the well-balanced covariates, these results support the validity of the subsequent logistic regression analysis for selecting candidate drugs.
Subsequent logistic regression analyses selected 8, 9, and 8 candidate drugs using Our (CR), BEHRT, and MED-BERT, respectively. Using diagnosis vector representations from each pretrained model enabled us to select multiple drugs that have been reported to exhibit protective associations in prior studies (Table \ref{table:evidence status}). Notably, the diagnosis vectors from Our (CR) model yielded the largest number of such drugs. Importantly, by leveraging representations learned through pre-training, the candidate drugs were derived solely from the data structure without relying on external knowledge (e.g., the literature). This demonstrates that drugs previously deemed efficacious can be rediscovered in a fully data-driven manner, indicating that the learned representations captured meaningful relationships between disease states and prescribing behavior. This also suggests the proposed method's potential to generate plausible hypotheses for previously unknown candidate drugs. Figure \ref{fig:dist_estimated_ors} shows the distribution of estimated ORs for all screened drugs. The distribution peaks at around OR = 1, indicating that most drugs exhibit no strong associations, while drugs with meeting our screening criteria for protective associations concentrated in the left tail. Although this distribution does not represent a strict null distribution, it shows that the selected candidate drugs occupy relatively extreme positions within the overall  distribution, even under conditions where treatment effects may be partially absorbed (as noted in Section \ref{candidate selection method}). This suggests that the proposed approach selectively extracts a limited subset of drugs rather than assigning uniform associations across all medications.

\subsection{Prioritization of Selected Candidates }
\begin{figure}[b]
    \centering
    \begin{minipage}[b]{0.45\linewidth}
        \centering
        \includegraphics[width=\linewidth]{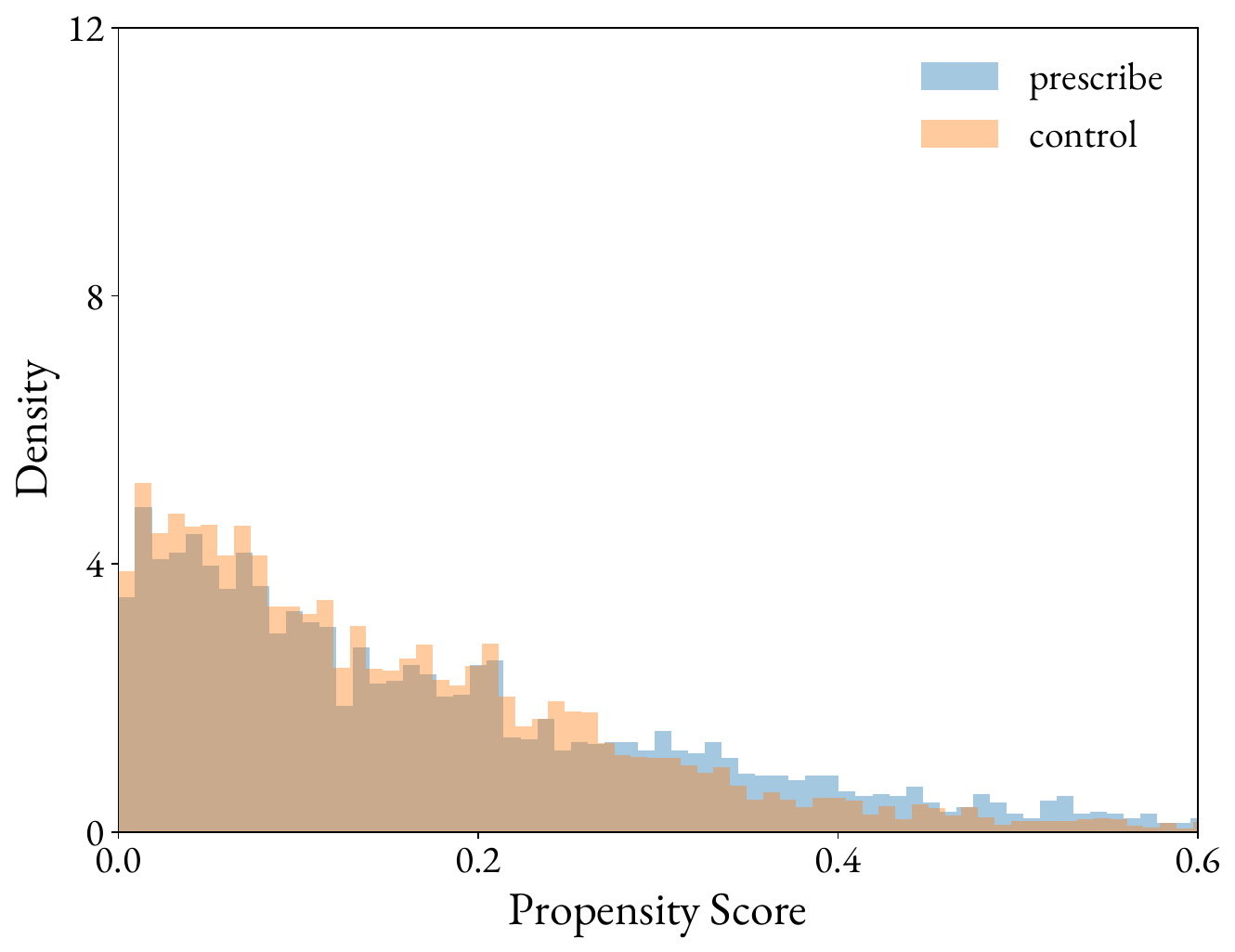}
        \subcaption{Without TARA (Original)} 
        \label{fig:candesartan propensity score dist without tara}
    \end{minipage}%
    \hfill
    \begin{minipage}[b]{0.45\linewidth}
        \centering
        \includegraphics[width=\linewidth]{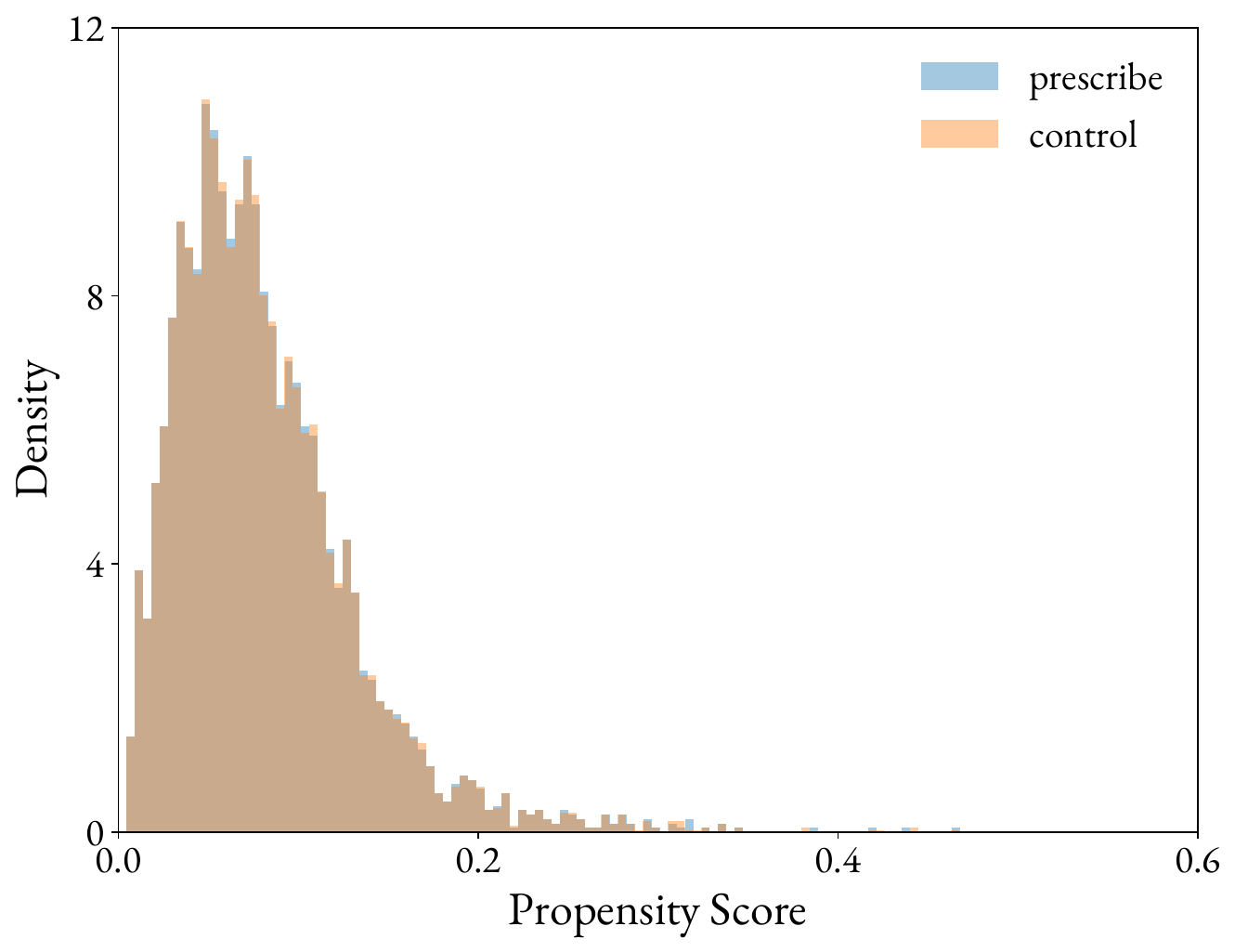}
        \subcaption{With TARA}
        \label{fig:candesartan propensity score dist with tara}
    \end{minipage}
    \caption{\textbf{Candesartan Propensity Score Distribution Change}}
    \label{fig:candesartan propensity score dist}
\end{figure}

Figure \ref{fig:candesartan propensity score dist} shows the propensity score distributions for candesartan using diagnosis vectors from Our (CR), with and without TARA. Doing so improved the overlap between treated and control groups, particularly in regions where the propensity score exceeded 0.2. This suggests that suppressing prescription-related information encoded in diagnosis vectors enables the construction of matched cohorts that more broadly reflect the original patient population, thereby improving the robustness of hypothesis prioritization.
Using diagnosis vectors generated by Our (CR) with TARA, we estimated the associations between regular prescription of candidate drugs and AD onset to prioritize candidate hypotheses. In this analysis, after controlling for multiple testing, four drugs—candesartan, vildagliptin, pitavastatin, and loxoprofen—were prioritized (FDR adjusted \textit{p} $<$ 0.05) and exhibited protective associations. In contrast, when using diagnosis vectors from the benchmark models, only loxoprofen for BEHRT and three drugs (candesartan, pitavastatin, and other ophthalmologicals) for MED-BERT met the same (for details on the estimated associations, see Table \ref{tab:all_associations}). For the candidate drugs, applying TARA to Our (CR) ensures an adequate propensity score overlap; similarly, a sufficient overlap was observed when using diagnosis vectors from BEHRT and MED-BERT. However, for a subset of candidate drugs, sufficient overlap could not be achieved, regardless of whether the diagnosis vectors were generated using Our (CR) with TARA or using benchmark representations from BEHRT or MED-BERT. These cases likely corresponded to drugs whose prescription likelihood can be readily inferred from diagnoses alone. For example, mecobalamin and platelet aggregation inhibitors excluding heparin are prescribed in 38\% and 43\% of cases for lumbar spinal canal stenosis and polyneuropathy (unspecified), respectively, making prescription status relatively easy to infer from diagnostic information. Consequently, even when prescription information is partially or fully masked, constructing matched cohorts that adequately reflect the original population is difficult. We further evaluated SMDs before and after matching for the covariates used in the propensity score estimation, namely diagnosis vectors, sex, and age, using diagnosis representations from Our (CR) with TARA, BEHRT and MED-BERT. Across all diagnostic representations, post-matching SMDs are below 0.1 for age and sex, and for more than 96\% of the diagnosis vector dimensions. This indicates that, regardless of the diagnostic representation employed, adequate covariate balance was achieved after matching within the covariate space considered in this study. Detailed post-matching covariate balance results, including all figures corresponding to diagnosis vectors from Our (CR) with TARA, are provided in Figure \ref{fig:appendix_smd_plots_propensity_score_matching}. Taken together, these results indicate that the differences among models were not primarily attributable to post-matching covariate balance itself, but rather due to differences in the range of patients that can be matched and in the structural properties of the resulting propensity score distributions. 
Table \ref{table:four smds of clinical indicators in propensity score matching} reports SMDs for clinical severity indicators for each candidate drug (results for drugs beyond the four highlighted are provided in Table \ref{table:appendix additonal smds of clinical indicators in propensity score matching}). Except for vildagliptin, clinical indicators remained well balanced after applying TARA, with improvements in propensity-score overlap also observed. These results suggest that, for most drugs, TARA enables robust hypothesis prioritization by improving propensity score overlap without substantially degrading balance on clinical severity indicators. 
Finally, ablation results (Table \ref{tab:tara_ablation}) confirm that protective signals persist without TARA but fail to reach statistical significance. The findings highlight that representation quality alone is insufficient for robust hypothesis prioritization; controlling prescription leakage via TARA is essential to align learned representations with this objective.
\begin{table}[h]
    \setcounter{table}{3} 
    \caption{\textbf{SMDs of clinical severity indicators in propensity score matching using Our (CR) diagnosis vector}}
    \label{table:four smds of clinical indicators in propensity score matching}
    \centering
    \resizebox{1.0 \linewidth}{!}{%
        \begin{tabular}{ccccccccc}
            \toprule
            & \multicolumn{2}{c}{vildagliptin}
            & \multicolumn{2}{c}{candesartan}
            & \multicolumn{2}{c}{loxoprofen}
            & \multicolumn{2}{c}{pitavastatin} \\
            \cmidrule(lr){2-3} \cmidrule(lr){4-5} \cmidrule(lr){6-7} \cmidrule(lr){8-9}
            & Original & TARA & Original & TARA & Original & TARA & Original & TARA \\
            \midrule\midrule 
            CCI & \textbf{0.094} & 0.14  
            & 0.04 & \textbf{0.022}  
            & \textbf{0.024} & 0.042  
            & 0.062 & \textbf{0.016} \\
            NDPM$^1$
            & \textbf{0.059} & 0.112
            & 0.045 & \textbf{0.006}
            & \textbf{0.067} & 0.069
            & 0.06 & \textbf{0.025} \\
            \bottomrule
        \end{tabular}
        }
        \flushleft \footnotesize{1) Number of distinct prescribed medicines identified by ATC codes per year.}
\end{table}

\section{Conclusion}
Our findings from this study should be interpreted strictly as hypothesis generation and prioritization results, rather than as evidence of causal effects. In this study, we proposed a unified pre-training framework that integrates the hierarchical structure of medical codes and the diagnosis–treatment interactions commonly present in medical claims data and electronic health records. The proposed methods---Hierarchical Sub-token Aggregation, Partial Masking, and Cross-Reference mechanism ---consistently outperformed existing BERT-based models and substantially improved prediction performance. A key contribution of this work is the demonstration of a practical hypothesis discovery workflow. This workflow effectively integrates two strengths: (1) our domain-informed pre-training model, which enables promising hypothesis generation by capturing complex diagnosis–treatment relationships; and (2) the Task-Adaptive Representation Approach, which ensures robust hypothesis prioritization by enabling analyses that broadly reflect the original patient population. Accordingly, candidate hypotheses with protective associations passing false discovery rate control were prioritized. Our screening workflow generates and prioritizes hypotheses based on observational signals.  Looking forward, further refining this framework will necessitate addressing inherent data-source biases and methodological limitations---a detailed discussion of which is provided in Appendix \ref{appendix limitations}. More broadly, this work highlights the importance of representation control in knowledge discovery workflows, where the optimal representation depends on the analysis objective. We believe this perspective is relevant beyond healthcare and can inform representation learning for hypothesis discovery from structured data. 

\section*{Impact Statement}
This work aims to advance Machine Learning. There are many potential societal consequences of our work, none which we feel must be specifically highlighted here.

\clearpage

\bibliography{reference}
\bibliographystyle{icml2026}

%%%%%%%%%%%%%%%%%%%%%%%%%%%%%%%%%%%%%%%%%%%%%%%%%%%%%%%%%%%%%%%%%%%%%%%%%%%%%%%
%%%%%%%%%%%%%%%%%%%%%%%%%%%%%%%%%%%%%%%%%%%%%%%%%%%%%%%%%%%%%%%%%%%%%%%%%%%%%%%
% APPENDIX
%%%%%%%%%%%%%%%%%%%%%%%%%%%%%%%%%%%%%%%%%%%%%%%%%%%%%%%%%%%%%%%%%%%%%%%%%%%%%%%
%%%%%%%%%%%%%%%%%%%%%%%%%%%%%%%%%%%%%%%%%%%%%%%%%%%%%%%%%%%%%%%%%%%%%%%%%%%%%%%
\newpage
\appendix
\onecolumn

\setcounter{table}{0}
\renewcommand{\thetable}{A\arabic{table}}
\setcounter{figure}{0}
\renewcommand{\thefigure}{A\arabic{figure}}
\section{Masking Strategies Supplement}
\label{appendix masking strategy}
\begin{figure*}[h]
    \raggedleft 
    \includegraphics[width=1.0\linewidth]{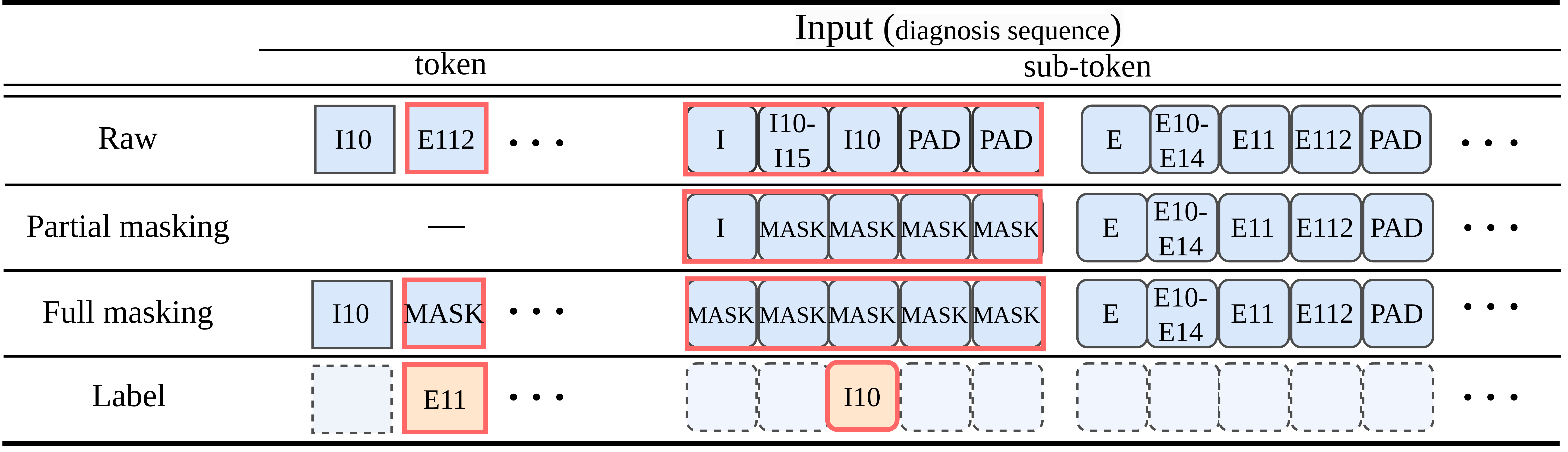}
    \caption{\textbf{Overview of the masking strategies.} Tokens selected as masking targets are outlined in red, and tokens corresponding to labels are highlighted in orange. In partial masking, the information for the first sub-token is provided to the model, whereas no information is provided in full masking. Regardless of whether the input sequence consists of tokens or sub-tokens, the third-level classification (the central sub-token) of the diagnosis token was used as the label.}
    \label{fig_masking_strategies}
\end{figure*}

\section{Our Proposed Pre-training Model Supplement}
\label{appendix archi supplement}
\begin{figure}[h!]
    \centering
    \includegraphics[width=0.94\linewidth]{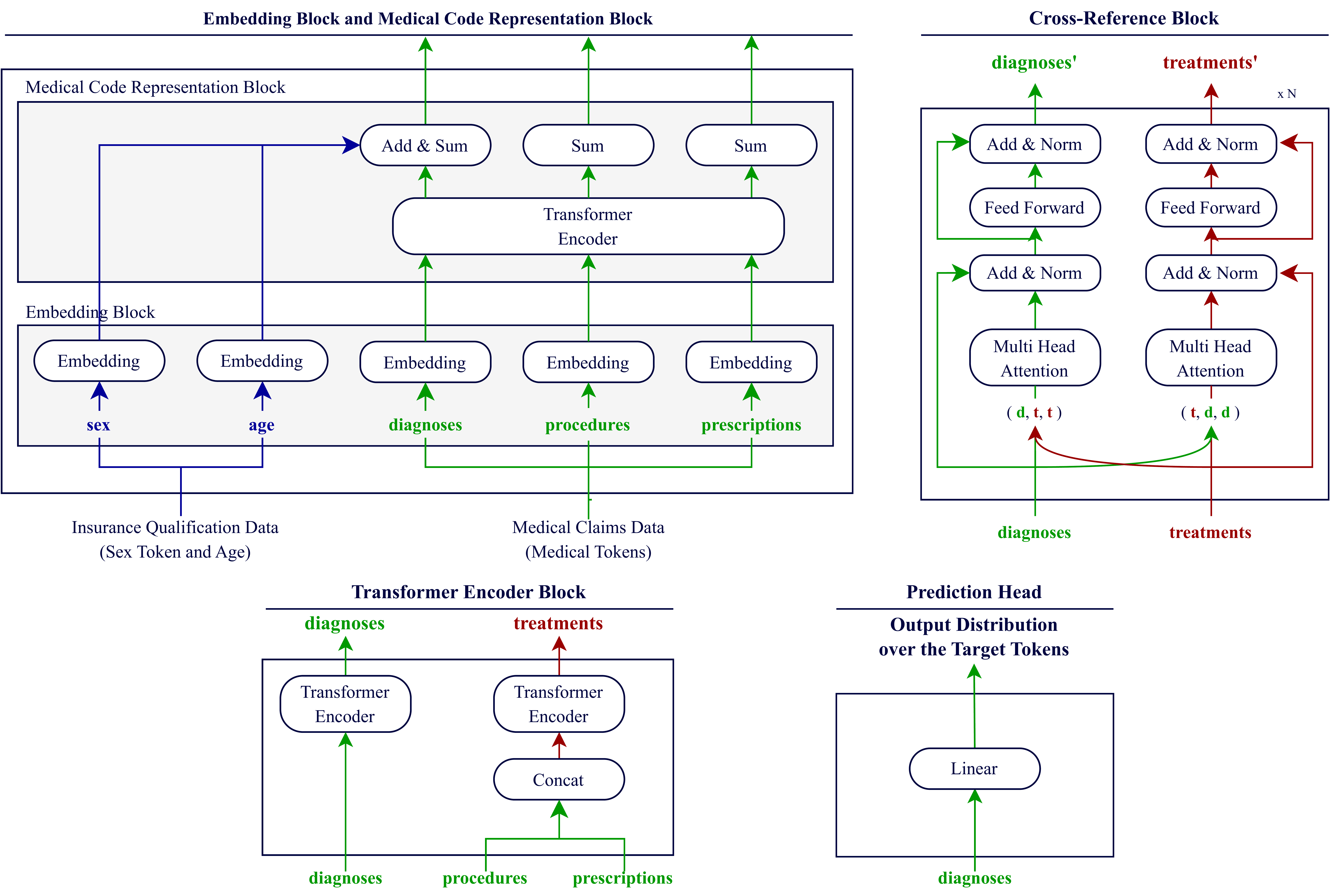}
    \smallskip
    \caption{\textbf{Blocks in our proposed pre-training model.}}
    \label{fig_blocks}
\end{figure}

\clearpage
\section{Details of Clinical Event Prediction Task}
\subsection{Our Fine-Tuning Model}\label{appendix fine-tuning model}
We developed a fine-tuning model composed of three blocks for clinical event prediction, including dementia and hospitalization. The input consists of the pre-training inputs spanning 12 months. These inputs are first processed by pretrained modules, followed by a clinical representation block, and finally a classification head that outputs event probabilities. \textbf{(i) Pretrained Modules:} This module reuses the pretrained components described in Section \ref{our pre-training model}, ranging from the embedding block to the CR block. Given 12 months of input features, the module outputs a pair of diagnosis and treatment vectors for each month. \textbf{(ii) Clinical Representation Block:} The diagnosis and treatment vectors produced by the pretrained modules are concatenated to form monthly clinical vectors that are then concatenated to construct a clinical sequence that encapsulates one year of diagnosis and treatment information. Standard positional encoding for Transformers \cite{NIPS2017_3f5ee243} is applied to this sequence and subsequently processed by a Transformer encoder to update the monthly clinical vectors. The updated 12-month vectors are then summed along the temporal dimension to produce a single annual clinical vector representing the patient’s one-year medical history. \textbf{(iii) Classification Head:} The annual clinical vector is input to a classification head consisting of a linear layer, L2 normalization, and a scale layer \cite{ranjan2017l2}. The classification head then outputs the probabilities of hospitalization and the onset of dementia.

\subsection{Experimental Setting}\label{appendix_clinical_predicton_task_experimental_setting} 
As downstream tasks, we evaluated the predictions for dementia onset and hospitalization. The pretrained models selected for fine-tuning were determined based on their performance in the pre-training task. To evaluate the effect of architectural differences (Self-Attention vs. Cross-Reference), we fine-tuned Our (CR) and Our (SA) that achieved the best pre-training performance with both HSA and PM applied. Additionally, to explicitly examine the incremental effects of HSA and PM, we conducted further ablation studies on CR by fine-tuning three variants: (i) a model without either HSA or PM, (ii) a model using HSA only, and (iii) a model using both HSA and PM. The baseline models (BEHRT and MED-BERT) were fine-tuned using their standard pretrained versions. To mitigate fine-tuning instability \cite{dodge2020finetuningpretrainedlanguagemodels, mosbach2020stability}, we followed prior research \cite{zhang2021revisiting} and reinitialized the parameters of the last two transformer layers before fine-tuning. All pretrained models and LGBM were trained and compared using predictions of dementia onset and hospitalization. We first held out 10\% of the dataset as a stratified test set. The remaining dataset was split using stratified 9-fold cross-validation for training and validation. After 10 epochs of training, the checkpoint with the highest validation ROC-AUC was selected. Evaluation was performed over nine folds, and the mean test performance was reported. We defined dementia onset using ICD-10 codes F00–F03 and G30.

\subsection{Results}\label{Appendix results of clinical event prediction task}The results of the clinical event prediction are summarized in Table \ref{results of clinical event prediction tasks}. For the onset of dementia, applying HSA and PM improved the prediction performance, and Our (CR) achieved the best results, with PR-AUC experiencing an improvement of over 10\%. In contrast, the effects of HSA and PM on hospitalization prediction were limited. This limitation may reflect the nature of the task: unlike dementia onset, hospitalization often depends on short-term and acute clinical events that are not fully captured in historical claims data. Therefore, the observed performance difference may stem from differences in the temporal characteristics of the target outcomes. In summary, these results suggest that the proposed model aligns with long-term dementia progression better than acute clinical events, indicating that the proposed approach captures gradually evolving conditions more effectively than short-term acute events.

\begin{table}[H]
    \centering
    \caption{\textbf{Results of Clinical Event Prediction tasks.}}
    \label{results of clinical event prediction tasks}
    \resizebox{0.63\linewidth}{!}{
    \begin{tabular}{cccccccc}
        \toprule[1.5pt]
        \multirow{2}{*}{Model} & \multirow{2}{*}{Proposed} & \multicolumn{3}{c}{Dementia Onset} & \multicolumn{3}{c}{Admission} \\
        \cmidrule(lr){3-5} \cmidrule(lr){6-8} 
         &  Methods & PR-AUC & ROC-AUC & MCC & PR-AUC & ROC-AUC & MCC \\
        \midrule\midrule 
        
        \multirow{3}{*}{Our (CR)} 
         & & 0.1949 & 0.8073 & 0.2450 & 0.5061 & 0.6856 & 0.2632 \\
         & HSA & 0.1998 & 0.8147 & \textbf{0.2652} & 0.5020 & 0.6859 & \textbf{0.2669} \\
         & HSA+PM & \textbf{0.2204} & \textbf{0.8164} & 0.2577 & 0.5043 & \textbf{0.6893} & 0.2610 \\
        \midrule
        
        Our (SA) & HSA+PM & 0.2049 & 0.8094 & 0.2459 & \textbf{0.5067} & 0.6843 & 0.2644 \\
        \midrule
        
        BEHRT & & 0.1765 & 0.8014 & 0.2449 & 0.4911 & 0.6799 & 0.2533 \\
        \midrule
        
        MED-BERT & & 0.1303 & 0.7177 & 0.1588 & 0.4767 & 0.6548 & 0.2100 \\
        \midrule
        
        LGBM & & 0.1874 & 0.8085 & 0.2329 & 0.4902 & 0.6820 & 0.2142 \\
        \bottomrule[1.5pt]
    \end{tabular}
    }
\end{table}

\clearpage
\section{Hyperparameters Supplement}
\label{appendix hyperparameter}
\begin{table*}[h]
\caption{\textbf{Hyperparameters for the pre-training models}}
\label{hyperparameter table}
\centering
\begin{small}
\resizebox{1.0\linewidth}{!}{
\begin{tabular}{llcccc}
\toprule[0.4mm]
\multirow[c]{2}{*}{Category} & \multirow[c]{2}{*}{Parameter} & \multirow[c]{2}{*}{\makecell[c]{Our Proposed Model\\( Our (CR) )}} & \multirow[c]{2}{*}{Our (SA)} & \multirow[c]{2}{*}{BEHRT} & \multirow[c]{2}{*}{MED-BERT}\\
 & &  &  \\ \midrule[0.1mm]\midrule[0.1mm]
Batch & batch size & 128 & 128 & 128 & 128 \\
\multirow[t]{2}{*}{Embedding Block} &  num embeddings & \num{20000} & \num{20000} & \num{11000} & \num{10000} \\
 &  embedding dim & 256 & 256 & 288 & 192 \\
 \multirow[t]{5}{*}{Medical Code Representation Block} & d\_model & 256 & 256 & - & -\\
 &  n\_heads & 8 & 8 & - & - \\
 &  dim\_feedforward & 1024 & 1024 & - & - \\
 &  num\_layers & 2 & 2 & - & - \\
 &  dropout & 0.1 & 0.1 & - & - \\
\multirow[t]{5}{*}{Transformer Encoder Block} & d\_model & 256 & 256 & 288 & 192\\
 &  n\_heads & 8 & 8 & 12 & 6 \\
 &  dim\_feedforward & 1024 & 1024 & 512 & 64 \\
 &  num\_layers & 2 & 6 & 6 & 6 \\
 &  dropout & 0.1 & 0.1 & 0.1 & 0.1 \\
\multirow[t]{5}{*}{Cross-Reference Block} & d\_model & 256 & - & - & -\\
 &  n\_heads & 8 & - & - & - \\
 &  dim\_feedforward & 1024 & - & - & - \\
 &  num\_layers & 4 & - & - & - \\
 &  dropout & 0.1 & - & - & - \\
\multirow[t]{2}{*}{Prediction Head} & in\_features & 256 & 256 & 288 & 192\\
 & out\_features & 9012 & 9012 & 9012 & 9012 \\
\multirow[t]{3}{*}{Loss function} & name & Cross Entropy Loss &  & \multirow{9}{*}{The same as Our (CR)} & \\
 & weight &  None &  &  &\\
 & reduction & mean &  &  &\\
 \multirow[t]{6}{*}{Optimizer} & name & adamW &  &  & \\
  & lr & 1e-4 &  &  & \\
  & beta1 & 0.9 &  &  & \\
  & beta2 & 0.999 &  &  & \\
  & eps & 1e-8 &  &  & \\
  & weight decay & 1e-4 &  &  & \\
\bottomrule[0.4mm]
\end{tabular}
}
\end{small}
\end{table*}

\section{Details of Benchmark Models}\label{appendix benchmark models}
We compared our approach with the following four baseline models: \textbf{(i) BEHRT:} BEHRT is a representative self-attention–based model built on the BERT architecture. It employs standard MLM for pre-training and uses diagnosis, age, segment, and positional tokens based on treatment-month indices as inputs. \textbf{(ii) MED-BERT:} MED-BERT is another self-attention-based model that has been widely cited as a benchmark \cite{pmlr-v158-pang21a, 10.1007/978-3-031-39539-0_7}. In the original MED-BERT, the NSP pre-training objective used in BERT is replaced with predicting the occurrence of prolonged hospital stays (7 days or more). However, since measuring the length of hospital stays was difficult in our dataset, we instead predicted the occurrence of hospital admission. The input consisted of diagnosis and positional tokens derived from treatment-month indices. \textbf{(iii) Our Self-Attention Only Model (Our (SA)):} This ablation model is designed to evaluate the effectiveness of the proposed cross-reference mechanism. While it is based on self-attention, unlike BEHRT and MED-BERT, it adopts our proposed pre-training methods, including hierarchical sub-token aggregation and partial masking. Further to diagnosis and age tokens, this model also incorporates procedure, prescription, and sex tokens that are not used in BEHRT or MED-BERT. \textbf{(iv) Light Gradient Boosting Machine (LGBM):} LGBM \cite{NIPS2017_6449f44a} is a gradient boosting framework known for its efficient training. Unlike the sequence-based models, LGBM uses tabular input features constructed from diagnoses, procedures, prescriptions, sex, and age.

\clearpage
\section{Input Features Supplement}\label{appendix_input_features}
\subsection{Monthly Input Features}\label{monthly input features}
The monthly input features included diagnosis codes (ICD-10), procedure codes (display codes), and prescription codes (ATC codes) recorded in monthly medical claims data, along with sex and age information obtained from insurance qualification data. These features were used for pretrained models employing the proposed methods: our CR-based model ((Our (CR) and Our(SA)). Each feature was represented as a token list, a sub-token list, an integer, a float, or a binary variable, depending on its type, as summarized in Table \ref{table_input_features}.\\
\textbf{Token lists:} To reduce redundancy, duplicate medical tokens within each month were removed to shorten the training time and reduce computational cost. Sex tokens and age were embedded in vectors and added to the diagnosis token representations by broadcasting them to match the length of the diagnosis token sequence, thereby ensuring compatibility across inputs.\\
\textbf{Sub-token lists:} Diagnosis, procedure, and prescription tokens recorded within a given month were further decomposed into sub-tokens according to their hierarchical structures. These were represented as multi-lists of sub-tokens defined as:
\begin{align*}
&\text{List}\big[\textstyle\prod_{i=1}^{5}\text{ICD-10}_{i}\big] = \big\{[A_1, \ldots, A_n] \mid A_n \in \textstyle\prod_{i=1}^{5} \text{ICD-10}_i \big\}\\
&\textit{where } \textstyle\prod_{i=1}^{5} \text{ICD-10}_i = \{(a_i, a_2, a_3, a_4, a_5) \mid a_i \in \text{ICD-10}_{i} \}, \text{ and } \text{ICD-10}_{i} \text{ is the } i\text{-th classification of ICD-10 code.}
\end{align*}
\subsection{Annual Input Features}
We generated three types of annual input features using the monthly input features constructed from medical claims data recorded over a 12-month period.\\
\textbf{Concatenated sequences:} Monthly input features for each patient were concatenated along the sequence direction. This representation is used as input for both pre-training and clinical event prediction in BEHRT and MED-BERT.\\
\textbf{Stacked sequences:} Monthly input features for each patient were stacked to form this feature. For months with no recorded medical claim data, a sequence consisting only of padded tokens was used as the monthly input feature. This feature was done to fine-tune the pretrained models based on the proposed methods.\\
\textbf{Tabular data:} Various tabular features were constructed depending on the task. For the clinical event prediction task, the input consisted of binary indicators for approximately 7,000 medical tokens, along with sex and age, and was used to train the LGBM. For the drug repositioning task, two distinct feature sets were used to estimate (i) prognostic and propensity scores, and (ii) ORs and \textit{p}-values for the associations between drugs and the onset of AD. For (i), sex, age, and diagnosis vectors generated from pretrained models were used as covariates, while the regular prescription status of the target drug was used as the explanatory variable for (ii).
\begin{table}[h]
\begin{minipage}{1.0\linewidth}
\caption{\textbf{Overview of input features.}}
\label{table_input_features}
\centering
\resizebox{1.0\linewidth}{!}{
\begin{tabular}{ccccccccccc}
\toprule[0.4mm]
\multirow[c]{2}{*}{Model} & \multirow[c]{2}{*}{Task} & \multirow[c]{2}{*}{Type} & \multirow[c]{2}{*}[0.7ex]{\makecell[c]{Max\\sequence\\length$^{*1}$}} & \multicolumn{7}{c}{Content}\\ \cmidrule[0.1mm]{5-11}
& & & & diagnosis & procedure & prescription & sex & age & position$^{*4}$ & segment$^{*5}$ \\ \midrule[0.1mm]\midrule[0.1mm]

\multirow[c]{2}{*}{Our (CR)} & Pre-training & Monthly & \multirow[c]{2}{*}{50 or 30$^{*2}$} & \multicolumn{3}{c}{\multirow[c]{2}{*}{multi-lists of sub-tokens$^{*3}$}} & \multirow[c]{2}{*}{lists of tokens} & float & \multirow[c]{2}{*}{-} & \multirow[c]{2}{*}{-}\\
& Clinical Event Prediction & Annual (stacked sequence) & & \multicolumn{3}{c}{} 
& & list of float & & \\ \cmidrule[0.1mm]{1-11}

\multirow[c]{2}{*}{Our (SA)} & Pre-training & Monthly & \multirow[c]{2}{*}{50 or 30$^{*2}$} & \multicolumn{3}{c}{\multirow[c]{2}{*}{multi-lists of sub-tokens$^{*3}$}} & \multirow[c]{2}{*}{lists of tokens} & float & \multirow[c]{2}{*}{-} & \multirow[c]{2}{*}{-}\\
& Clinical Event Prediction & Annual (stacked sequence)
& & \multicolumn{3}{c}{}  & & list of float & & \\
\cmidrule[0.1mm]{1-11}

\multirow[c]{2}{*}{BEHRT} & Pre-training & \multirow[c]{2}{*}{\makecell[c]{Annual\\(concatenated sequence)}} & \multirow[c]{2}{*}{312$^{*6}$} & \multirow[c]{2}{*}{list of tokens} & \multirow[c]{2}{*}{-} & \multirow[c]{2}{*}{-} & \multirow[c]{2}{*}{-} & \multicolumn{3}{c}{\multirow[c]{2}{*}{\makecell{list of tokens\\(age, position, segment)}}} \\
& \multirow[c]{1}{*}{Clinical Event Prediction} & & & & & & & \\ \cmidrule[0.1mm]{1-11}

\multirow[c]{2}{*}{MED-BERT} & Pre-training & \multirow[c]{2}{*}{\makecell[c]{Annual\\(concatenated sequence)}} & \multirow[c]{2}{*}{300} & \multirow[c]{2}{*}{list of tokens} & \multirow[c]{2}{*}{-} & \multirow[c]{2}{*}{-} & \multirow[c]{2}{*}{-} & \multirow[c]{2}{*}{-} & \multirow[c]{2}{*}{\makecell{list of tokens}} & \multirow[c]{2}{*}{-} \\
& \multirow[c]{1}{*}{Clinical Event Prediction}& & & & & & & & & \\ \cmidrule[0.1mm]{1-11}

\multirow[c]{3}{*}{\makecell[c]{LGBM$^{*7}$\\CBPS\\Logistic Regression}} & Clinical Event Prediction & Annual (tabular data) & - & \multicolumn{3}{c}{\makecell{bool\\(diagnosis, procedure, prescription)}} & bool & float & - & - \\ \cmidrule[0.1mm]{2-11}
& \multirow[c]{1}{*}{Drug repositioning$^{*8}$} & \multirow[c]{1}{*}{Annual (tabular data)} & \multirow[c]{1}{*}{-} & \multirow[c]{1}{*}{float} & - & \multirow[c]{1}{*}{bool} & \multirow[c]{1}{*}{bool} & \multirow[c]{1}{*}{int} & - & -\\

\bottomrule[0.4mm]
\end{tabular}
}

\begin{flushleft}
\scriptsize
1) If the sequence length exceeded the max sequence length, it was truncated to the max sequence length. \\
2) The max sequence length for prescriptions was set to 30, but to 50 for other tokens. \\
3) For Our (CR) and Our (SA), when our proposed methods were not applied, diagnosis, procedure, and prescription were represented as  lists of tokens, the same as in BEHRT and MED-BERT. \\
4) Temporal information represents the order of monthly medical claims recorded within a year. For example, if claims were recorded in March, May, and November of 2017, their positions would be 0, 1, and 2, respectively.\\
5) Tokens correspond to the parity (even or odd) of the positions.\\
6) The max sequence length for diagnosis, age, and segment tokens was 300. Only BEHRT utilized CLS and SEP tokens, which added up to 12 additional tokens (one per month), resulting in a total max sequence length of 312.\\7) LGBM is used for both tasks, while CBPS and Logistic Regression are used only for the drug repositioning task. 
\end{flushleft}
\end{minipage}
\end{table}

\clearpage
\section{Dataset Supplement}
\label{appendix datasets}
We used a database containing medical claims and insurance qualification data collected from multiple insurers operated by local governments in Japan. The source database contained medical claims data for 5,506,645 individuals and insurance qualification data for 5,647,546 individuals.\\
 \textit{\textbf{Dataset Construction and Splitting:}} To prevent data leakage and ensure fair evaluation, we first partitioned individuals with claims records into two mutually exclusive groups: (A) the Pre-training and Drug Repositioning group (80\%, \textit{n}=4,405,316) and (B) the Clinical Event Prediction group (20\%, \textit{n=}1,101,329). Across all tasks, individuals were included only if they met two common criteria: (i) their records contained the medical tokens required for cross-attention computation (at least one recorded diagnosis and at least one procedure or prescription token), and (ii) sex and age information were available in the insurance qualification data. We constructed task-specific datasets through a combination of random sampling and the application of these common and additional criteria, ensuring no overlap between the pre-training and downstream datasets. The procedures are detailed below:\\
 \textbf{(1) Pre-training Task Dataset}: Among the individuals in Group A, we identified 1,765,369 individuals with medical claims records between January and December 2017 who satisfied the common criteria. From these 1,765,369 individuals, we randomly sampled 50,000 individuals to construct the pre-training dataset. This dataset contains approximately 440,000 medical claims and 1.6 million medical tokens.\\
\textbf{(2) Clinical Event Prediction Task Dataset}: From Group B, we extracted 293,368 individuals for dementia onset prediction and 302,583 for hospitalization prediction who satisfied the following additional criteria and for whom we were able to learn about clinical events occurring between January 2018 and December 2020. Finally, 20,000 individuals were randomly sampled from each group to construct the datasets for the Clinical Event Prediction Tasks. The additional inclusion criteria were as follows: for dementia onset prediction, no prior history of dementia between January and December 2017; for hospitalization prediction, no hospitalization record in December 2017.  For both groups, we required continuous insurance enrollment from January 2017 through December 2020.\\
\textbf{(3) Drug Repositioning Task Dataset:} From group A, we randomly sampled an additional 100,000 individuals, ensuring no overlap with the pretraining dataset. We then extracted 80,308 individuals who satisfied the common and the following additional criteria. Additional inclusion criteria were as follows: 1) No prior history of dementia, including Alzheimer’s disease (AD), during the two-year period from January 2017 to December 2018, which corresponds to the diagnosis vector extraction and the regular prescription assessment period. 2) For individuals without AD onset, continuous insurance enrollment was required from the diagnosis extraction period through the outcome observation period (January 2017 to December 2020). For individuals with AD onset, continuous insurance enrollment was required from the diagnosis vector extraction period through the regular prescription assessment period (December 2017 to December 2018).\\
\textit{\textbf{Rationale for Sampling:}} We developed and evaluated the model using randomly sampled subsets drawn from a population of over 5 million individuals. This sampling strategy served two primary purposes. First, it allowed us to demonstrate the data efficiency of the proposed methods (HSA, PM, and CR), showing that robust representations can be learned from relatively small training sets by effectively incorporating domain-specific knowledge, even under computational resource constraints. Second, the use of subsets serves as a  proof of concept to validate research hypotheses efficiently, rather than focusing on maximizing scale-based predictive performance.  It also facilitates rapid experimental cycles before proceeding to computationally intensive training on the large-scale population.

\begin{table}[htbp]
\begin{minipage}{1.0\linewidth}
\caption{\textbf{Overview of the datasets used for Our (CR) model across the four tasks}}
\label{dataset table}
\centering
\begin{small}
\begin{tabular}{lccccc}
\toprule[0.4mm]
 & \multirow[c]{2}{*}{\makecell[c]{Summary\\Statistic}} & \multirow{2}{*}{Pre-training} & \multicolumn{2}{c}{Clinical Event Prediction} & \multirow[c]{2}{*}{\makecell[c]{Drug\\Repositioning}} \\ \cmidrule[0.1mm]{4-5}
 & &  & Dementia onset& Hospitalization &  \\ \midrule[0.1mm]\midrule[0.1mm]
Male (binary)    & - & 42.7\% &  42.0\%  & 41.4\% & 42.1\%\\
\multirow[t]{2}{*}{Age$^1$ (years)}  & mean & 74.4 & 73.6 & 74.2 & 73.5\\
    & std & 8.6 &  7.9 & 8.1 & 7.9\\
\multirow[t]{2}{*}{Number of unique medical tokens per individual$^2$}& mean &  69.8 &  67.0 & 67.4 & 67.0\\
& std & 46.6 & 42.5 & 42.5 & 41.6\\
\multirow[t]{2}{*}{Months with medical visits} & mean & 8.7 &  9.2 & 9.3 & 9.2\\
 & std & 3.6 & 3.5 & 3.4 & 3.5\\
Incidence of dementia onset& - & - &  6.2\% & - & -\\
Incidence of hospitalization & - & - &  - & 31.6\% & -\\
Incidence of alzheimer's disease onset & - & - &  - & - & 2.8\%\\
\bottomrule[0.4mm]
\end{tabular}
\begin{flushleft}
\footnotesize
1) Age was calculated as of 31 December 2017. 2) Medical tokens include diagnosis tokens, procedure tokens, prescription tokens.
\end{flushleft}
\end{small}
\end{minipage}
\end{table}

\clearpage
\section{Visualization of Disease Embedding}
\label{appendix disease embedding}
\begin{figure*}[h!]
    \centering

    \begin{subfigure}[b]{1.0\linewidth} 
        \centering
        \includegraphics[width=1.0\linewidth]{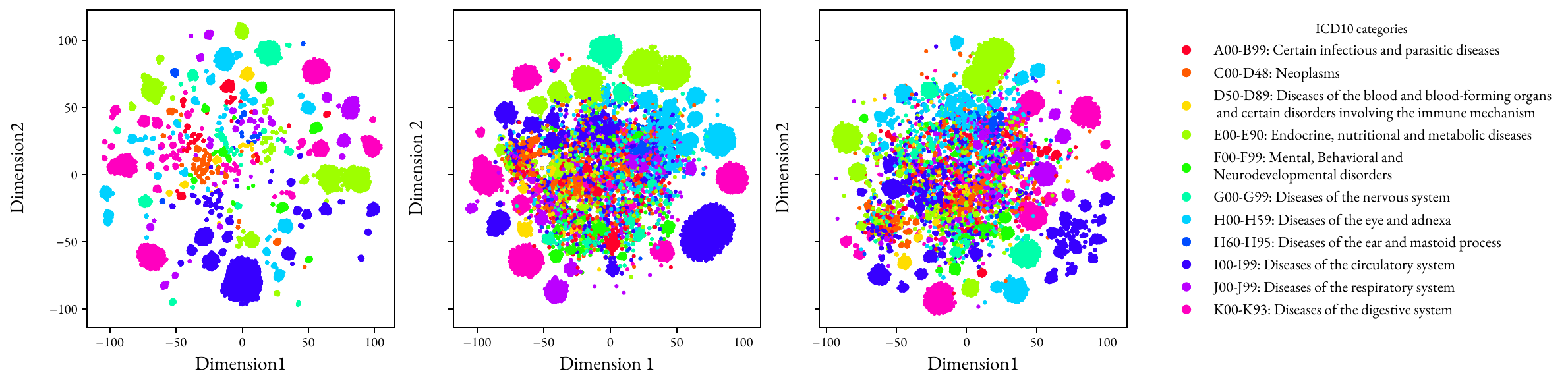}
        \caption{\textbf{Visualization of the first-level classifications: A--K.} From left to right, plots represent Our (CR), BEHRT, and MED-BERT models, respectively. In the benchmark models (BEHRT, MED-BERT), diagnosis tokens from different first-level classifications were scattered near the center of the plots. In contrast, such scattering was not observed in Our (CR).}
        \label{appendix tsne_a-k}
    \end{subfigure}

    \vspace{0.1in}

    \begin{subfigure}[b]{1.0\linewidth}  
        \centering
        \includegraphics[width=1.0\linewidth]{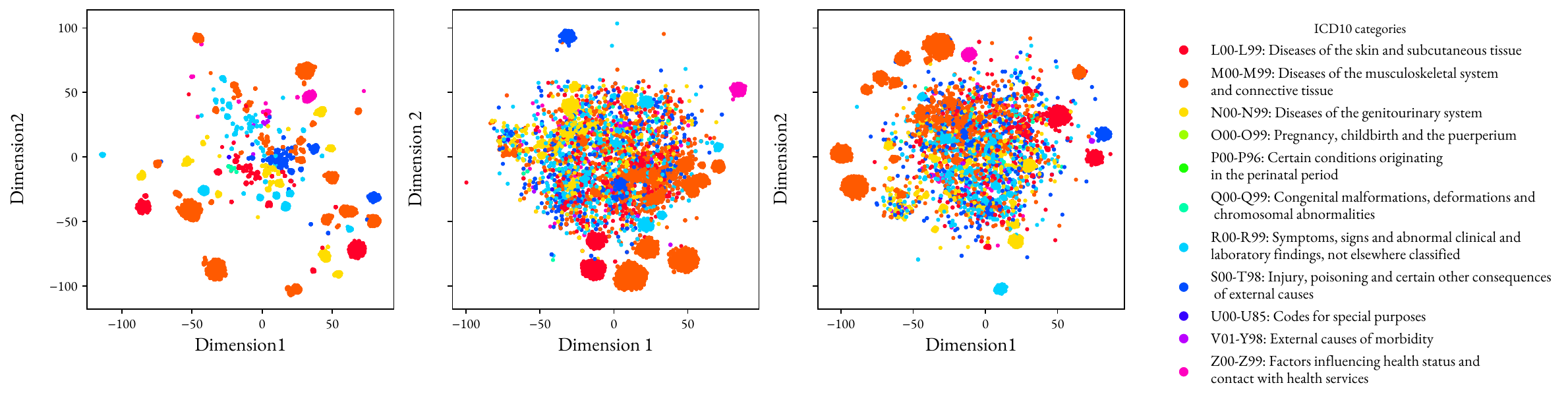}
        \caption{\textbf{Visualization of the first-level classifications: L--Z.} From left to right, plots represent Our (CR), BEHRT, and MED-BERT models, respectively. In the benchmark models (BEHRT, MED-BERT), diagnosis tokens from different first-level classifications were scattered near the center of the plots. In contrast, such scattering was not observed in Our (CR).}
        \label{tsne_l-z}
    \end{subfigure}

    \vspace{0.1in}

    \begin{subfigure}[b]{1.0\linewidth}  
        \centering
        \includegraphics[width=1.0\linewidth]{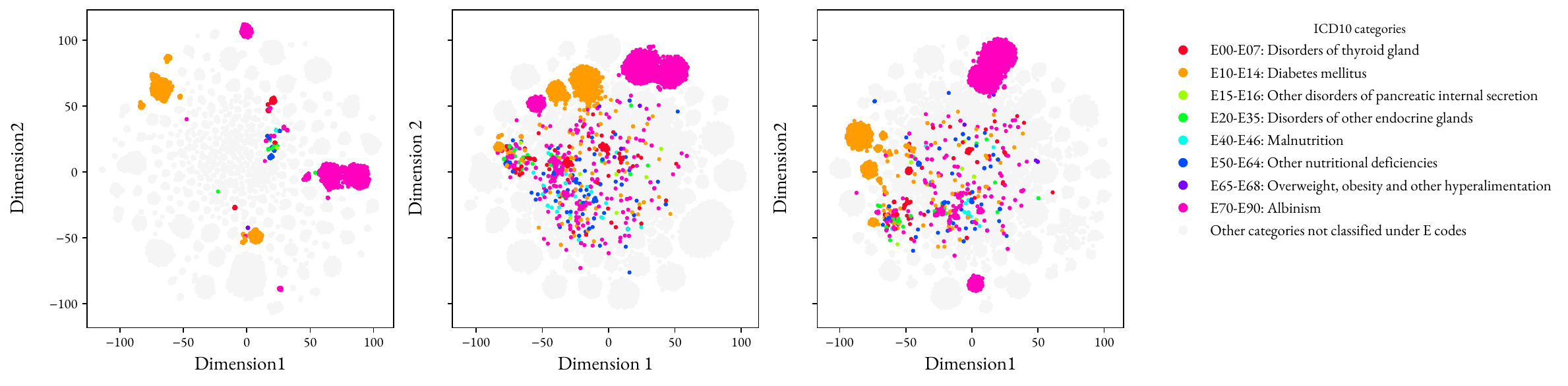}
        \caption{\textbf{Visualization of the disease embedding for the eight second-level classifications under first-level classification "E" }(endocrine, nutritional, and metabolic diseases). From left to right, plots represent Our (CR), BEHRT, and MED-BERT models, respectively. Our (CR) generated better feature representations than the benchmark models. For example, diagnosis tokens in second-level classifications such as E15–E16 (light green), E20–E35 (green), and E50–64 (blue) failed to form clusters and were scattered in the benchmark models. However, Our (CR) successfully formed clusters for these classifications.}
        \label{tsne_secondary}
    \end{subfigure}
    \caption{\textbf{Visualization of the Disease Embedding.}}
    \label{combined_tsne}
\end{figure*}

\clearpage
\section{Summary of Literature Review for Candidate drugs}
\label{appendix: summary of literature review}

\begin{enumerate}[label=\textbf{\arabic*.}, leftmargin=*, parsep=1em, itemsep=0.3em, parsep=0.2em]
\item \textbf{Drugs with Reported Protective Associations in Humans}\\
\textbf{Candesartan:} Large-scale retrospective cohort studies have reported that users of angiotensin II receptor blockers exhibit a 20–30\% lower risk of AD \cite{lundin2024association}.\\
\textbf{Pitavastatin:} Systematic reviews and meta-analyses have suggested that statin use, including pitavastatin, may be associated with a reduced risk of AD onset \cite{westphal2025statin}.\\
\textbf{Alendronic Acid:} Population-based observational studies have reported a lower risk of AD and all-cause dementia among users of bisphosphonates \cite{sing2025bisphosphonates}, including alendronic acid, compared with untreated individuals or users of other anti-osteoporosis drugs.
\item \textbf{Drugs with Protective Associations Not Established in Humans}
\begin{enumerate}[label=\textbf{\theenumi\arabic*}, leftmargin=1em, itemsep=0.2em, parsep=0.2em]
            \item \textbf{Protective Association Suggested by Preclinical or Indirect Evidence}\\
            \textbf{Vildagliptin:} Preclinical studies in rat suggest that vildagliptin has neuroprotective effects and may alleviate cognitive deficits \cite{ma2018vildagliptin}; however, consistent evidence supporting a protective association with AD prevention in large-scale human studies is lacking.\\
            \textbf{Teneligliptin:} Preclinical studies in mouse indicate that teneligliptin may attenuate diabetes-related cognitive impairment through anti-inflammatory mechanisms \cite{wang2023teneligliptin}, but evidence linking teneligliptin to AD prevention in human populations is currently lacking.\\
            \textbf{Irbesartan and Amlodipine:} To our knowledge, there is no direct evidence linking the combined use of irbesartan and amlodipine to AD onset in human populations. In contrast, irbesartan monotherapy has been reported to be associated with a reduced risk of AD and related dementia \cite{lundin2024association} \\
            \textbf{Loxoprofen:} Protective associations with AD have been reported for certain nonsteroidal anti-inflammatory drugs (NSAIDs) \cite{vom2025long}; however, loxoprofen itself has not been directly examined in that studies.\\
            \textbf{Ketoprofen:} While protective associations against AD have been reported for oral NSAIDs, including ketoprofen \cite{vom2025long}, the ketoprofen analyzed in the present study is a topical agent. Consequently, no protective association was found for topical ketoprofen.
            \item \textbf{No Protective Association Reported or Observed}\\
            Mecobalamin, Platelet Aggregation Inhibitors excluding Heparin, Amlodipine, Adenosine, Pregabalinum, Other Ophthalmologicals
        \end{enumerate}
\end{enumerate}

\section{Estimated Associations for Selected Candidate Drug Prioritization}
\label{appendix:estimated_asscociation_prioritization}
\begin{table}[h]
    \centering
    \small
    \setlength{\tabcolsep}{3.5pt}
    \caption{\textbf{Estimated associations between regular prescription of candidate drugs and the onset of AD.}}
    \label{tab:all_associations}
    
    \begin{subtable}{1.0\linewidth}
        \caption{Using diagnosis vectors from \textbf{Our (CR)}.}
        \label{tab:all_associations_our_CR}
        \centering
        \footnotesize{
            \begin{tabular}{lcccccccc}
                \toprule
                 & Vildagliptin & Candesartan & Loxoprofen & Pitavastatin & \makecell{Irbesartan and \\ Amlodipine} & PAIH$*$ & Mecobalamin & \makecell{Alendronic\\Acid} \\ \midrule
                odds ratio       & 0.466 & 0.698 & 0.768 & 0.746 & 1.000 & 0.946 & 0.849 & 0.834 \\
                p-value          & 0.009 & 0.006 & 0.003 & 0.017 & 1.000 & 0.630 & 0.156 & 0.279 \\
                adjusted p-value   & \textbf{0.024} & \textbf{0.024} & \textbf{0.024} & \textbf{0.034} & 1.000 & 0.72 & 0.250 & 0.372 \\
                \bottomrule
            \end{tabular}%
        }
    \end{subtable}% 
    \vspace{1mm}
    
    \begin{subtable}{1.0\linewidth}
        \caption{Using diagnosis vectors from \textbf{BEHRT}.}
        \centering
        \footnotesize{
            \begin{tabular}{lccccccccc}
                \toprule
                & Vildagliptin & Candesartan & Loxoprofen & Pitavastatin & Amlodipine & \makecell{OO$*$} & Mecobalamin & Adenosine & Ketoprofen \\ \midrule
                odds ratio       & 0.580 & 0.756 & 0.848 & 0.806 & 0.894 & 0.820 & 0.775 & 0.818 & 0.848 \\
                p-value          & 0.067 & 0.036 & 0.003 & 0.081 & 0.048 & 0.067 & 0.024 & 0.447 & 0.016\\
                adjusted p-value  &  0.086 & 0.081 & \textbf{0.027} & 0.091 & 0.086 & 0.086 & 0.072 & 0.447 & 0.072
                \\ \bottomrule
            \end{tabular}
    }
    \end{subtable}%  
    
    \vspace{1mm}

    \begin{subtable}{1.0\linewidth}
        \caption{Using diagnosis vectors from \textbf{MED-BERT}.}
        \centering
        \footnotesize{
            \begin{tabular}{lcccccccc}
                \toprule
                & Vildagliptin & Teneligliptin & Candesartan & Pitavastatin & PAIH$*$ & OO$*$ & Mecobalamin & Pregabalinum \\ \midrule
                odds ratio       & 0.569 & 0.621 & 0.613 & 0.711 & 0.790 & 0.761 & 0.792 & 0.801 \\
                p-value          & 0.057 & 0.091 & 0.000 & 0.005 & 0.032 & 0.011 & 0.039 & 0.166 \\
                adjusted p-value          & 0.076 & 0.104 & \textbf{0.000} & \textbf{0.02} & 0.062 & \textbf{0.029} & 0.062 & 0.166\\
                \bottomrule
            \end{tabular}%
    }
    \end{subtable}
\end{table}
\footnotesize (*) PAIH and OO are abbreviations for Platelet Aggregation Inhibitors excluding Heparin and Other Ophthalmologicals, respectively.

\section{Balance Assessment of Covariates and Clinical Severity Indicators after Propensity Score Matching}
\label{appendix_propensity_score_matching}
\begin{figure}[h]
    \centering
    \includegraphics[width=1.0\linewidth]{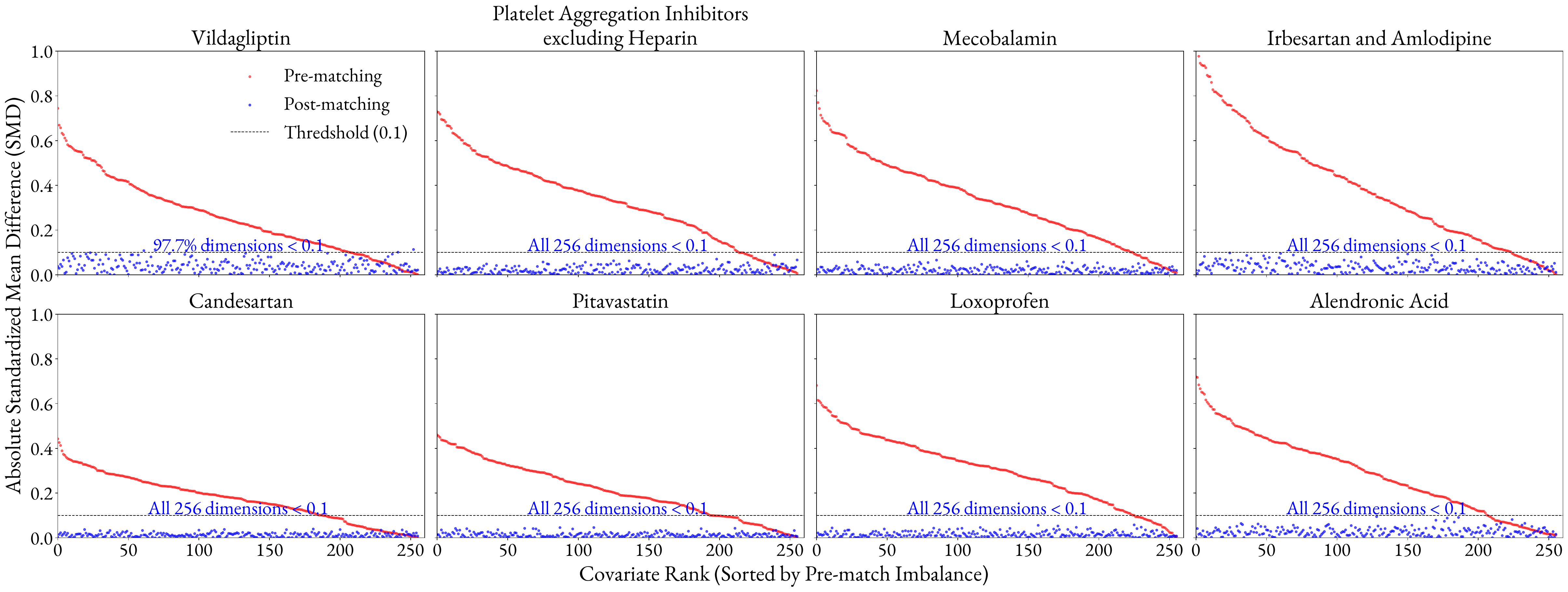}
    \caption{\textbf{SMD plots of Our (CR) Diagnosis Vector in Propensity Score Matching.}}
    \label{fig:appendix_smd_plots_propensity_score_matching}
\end{figure}

\begin{table}[htbp]
    \caption{\textbf{SMDs of clinical severity indicators in propensity score matching using Our (CR) diagnosis vector.} Results for Vildagliptin to Pitavastatin are reproduced.}
    \label{table:appendix smds of clinical indicators in propensity score matching}
    \centering
    \begin{subtable}{0.7\linewidth}
        \centering
        \caption{Reproduced results}
        \resizebox{\linewidth}{!}{%
        \begin{tabular}{ccccccccc}
            \toprule
            & \multicolumn{2}{c}{Vildagliptin}
            & \multicolumn{2}{c}{Candesartan}
            & \multicolumn{2}{c}{Loxoprofen}
            & \multicolumn{2}{c}{Pitavastatin} \\
            \cmidrule(lr){2-3} \cmidrule(lr){4-5} \cmidrule(lr){6-7} \cmidrule(lr){8-9}
            & Original & TARA & Original & TARA & Original & TARA & Original & TARA \\
            \midrule\midrule
            CCI & \textbf{0.094} & 0.14
            & 0.04 & \textbf{0.022}
            & \textbf{0.024} & 0.042
            & 0.062 & \textbf{0.016} \\
            NDPM$^1$
            & \textbf{0.059} & 0.112
            & 0.045 & \textbf{0.006}
            & \textbf{0.067} & 0.069
            & 0.06 & \textbf{0.025} \\
            \bottomrule
        \end{tabular}%
        }
    \end{subtable}
    
    \vspace{2mm} 
    
    \begin{subtable}{\linewidth}
        \centering
        \caption{Additional results}
        \label{table:appendix additonal smds of clinical indicators in propensity score matching}
        \resizebox{0.9\linewidth}{!}{%
        \begin{tabular}{ccccccccc}
            \toprule
            & \multicolumn{2}{c}{\makecell{Irbesartan and Amlodipine}}
            & \multicolumn{2}{c}{\makecell{Platelet Aggregation Inhibitors\\excluding Heparin}}
            & \multicolumn{2}{c}{Mecobalamin}
            & \multicolumn{2}{c}{Alendronic Acid}\\
            \cmidrule(lr){2-3} \cmidrule(lr){4-5} \cmidrule(lr){6-7} \cmidrule(lr){8-9}
            & Original & TARA & Original & TARA & Original & TARA & Original & TARA\\
            \midrule\midrule
            CCI & 0.031 & \textbf{0.014}
            & 0.018 & \textbf{0.002}
            & 0.019 & \textbf{0.012}
            & \textbf{0.023} & \textbf{0.03} \\
            NDPM$^1$
            & 0.121 & \textbf{0.065}
            & \textbf{0.034} & 0.09
            & \textbf{0.066} & 0.079
            & \textbf{0.014} & 0.035\\
            \bottomrule
        \end{tabular}%
        }
    \end{subtable}
    \flushleft \small{1) Number of distinct prescribed medicines identified by ATC codes per year.} 
\end{table}

\clearpage
\section{Ablation Study: Drug Prioritization without TARA}
\label{appendix:synergy_representation_and_tara}

To assess the respective contributions of representation quality and TARA to drug prioritization, we compared Our (CR) with and without applying TARA. Table \ref{tab:tara_ablation} reports the estimated odds ratios and p-values for all candidate drugs under both settings. Without TARA, no candidate drug reached statistical significance after FDR correction (all adjusted \textit{p} $>$ 0.05). In contrast, with TARA, four drugs were prioritized at FDR-adjusted \textit{p} $<$ 0.05. However, all odds ratios remained below 1.0 regardless of whether TARA was applied, indicating that the protective signals captured by Our (CR) persist in the learned representations.\\
The loss of statistical significance can be attributed to degraded overlap between the propensity score distributions of the treated and control groups. Because Our (CR) learns diagnosis–treatment interactions during pre-training, prescription information of target drug becomes over-encoded in the diagnosis vectors when TARA is not applied. This makes it difficult to construct well-overlapping matched groups, as illustrated by the propensity score distributions in Figure \ref{fig:candesartan propensity score dist without tara}. By suppressing this prescription leakage, TARA improves the overlap between treated and control groups (Figure \ref{fig:candesartan propensity score dist with tara}), enabling robust hypothesis prioritization. These results confirm that the rich representations of Our (CR) and TARA function synergistically: the former captures clinically meaningful protective signals, while the latter ensures the distributional overlap required to detect them.

\begin{table}[h]
    \centering
    \small
    \setlength{\tabcolsep}{3.5pt}
    \caption{\textbf{Candidate drug prioritization results for Our (CR) with and without TARA.}}
    \label{tab:tara_ablation}

    \begin{subtable}{\linewidth}
        \caption{\textbf{without} TARA}
        \centering
            \begin{tabular}{lcccccccc}
                \toprule
                 & Vildagliptin & Candesartan & Loxoprofen & Pitavastatin & \makecell{Irbesartan and \\ Amlodipine} & PAIH* & Mecobalamin & \makecell{Alendronic\\Acid} \\ \midrule
                odds ratio       & 0.499 & 0.736 & 0.826 & 0.791 & 0.806 & 0.774 & 0.849 & 0.775 \\
                p-value          & 0.018 & 0.025 & 0.033 & 0.084 & 0.316 & 0.054 & 0.228 & 0.126 \\
                adjusted p-value          & 0.088 & 0.088 & 0.088 & 0.134 & 0.316 & 0.108 & 0.261 & 0.168 \\
                \bottomrule
            \end{tabular}%
    \end{subtable}   

    \vspace{1mm}

    \begin{subtable}{\linewidth}
        \caption{with TARA (reproduced from Table \ref{tab:all_associations_our_CR})}
        \centering
            \begin{tabular}{lcccccccc}
                \toprule
                 & Vildagliptin & Candesartan & Loxoprofen & Pitavastatin & \makecell{Irbesartan and \\ Amlodipine} & PAIH* & Mecobalamin & \makecell{Alendronic\\Acid} \\ \midrule
                odds ratio       & 0.466 & 0.698 & 0.768 & 0.746 & 1.000 & 0.946 & 0.849 & 0.834 \\
                p-value          & 0.009 & 0.006 & 0.003 & 0.017 & 1.000 & 0.630 & 0.156 & 0.279 \\
                adjusted p-value          & \textbf{0.024} & \textbf{0.024} & \textbf{0.024} & \textbf{0.034} & 1.000 & 0.72 & 0.250 & 0.372 \\
                \bottomrule
            \end{tabular}%
    \end{subtable}
\begin{flushleft}
\footnotesize (*) PAIH is an abbreviation for Platelet Aggregation Inhibitors excluding Heparin.
\end{flushleft}
\end{table}

\section{Limitations}
\label{appendix limitations}
This study has several limitations related to both the data and the proposed methodology. First, the data source carries inherent limitations. Medical claims data collected for billing purposes may include diagnosis codes that do not fully reflect the patients’ true clinical conditions, potentially affecting the accuracy of training labels and learned vector representations. Furthermore, in claims data, diagnostic codes often appear only after a series of clinical events such as tests or prescriptions. Consequently, the prediction task may partially capture signals associated with ongoing disease processes rather than true early prediction—a limitation particularly relevant for dementia, where diagnostic coding may occur at relatively advanced stages. Moreover, as we used medical claims from insurers for the elderly in Japan, the study population is biased toward elderly individuals, meaning that the generalizability of our findings to populations with different genetic backgrounds or healthcare systems is unclear. Second, there are limitations arising from sampling. Due to computational constraints, we conducted our analyses using a subset of the full dataset. Consequently, the model may not have sufficiently learned informative representations for rare diseases or infrequently prescribed drugs. Third, there are inherent limitations associated with the exploratory, hypothesis discovery-oriented nature of the drug repositioning analysis. This analysis primarily sought to generate and prioritize hypotheses based on observational signals, rather than to provide confirmatory or causal evidence. Consequently, the generated hypotheses may be affected by residual biases. For example, unobserved confounders not captured in claims data, such as laboratory test results or socioeconomic factors, may have influenced the observed signals. Furthermore, for drugs whose prescribing can be inferred from diagnostic information, it may be difficult to construct matched cohorts that broadly reflect the pre-matching population, potentially limiting the validity of hypothesis prioritization to specific subpopulations. Moreover, because the analysis emphasized comparisons between highly similar groups and used strict caliper-based matching, the resulting hypotheses were primarily informed by matched cohorts; and may thus not fully represent the entire treated population. Finally, it is important to note that the candidate drugs identified in this study include several agents whose protective associations have been reported in independent studies across different populations and study designs. While this consistency with external evidence provides indirect support for the broader applicability of the learned representations, formal external validation remains an important direction for future work.
%%%%%%%%%%%%%%%%%%%%%%%%%%%%%%%%%%%%%%%%%%%%%%%%%%%%%%%%%%%%%%%%%%%%%%%%%%%%%%%
%%%%%%%%%%%%%%%%%%%%%%%%%%%%%%%%%%%%%%%%%%%%%%%%%%%%%%%%%%%%%%%%%%%%%%%%%%%%%%%

\end{document}